\pdfoutput=1

\documentclass[11pt]{article}

\usepackage{acl}

\usepackage{times}
\usepackage{latexsym}
\usepackage{enumitem}

\usepackage[T1]{fontenc}

\usepackage[utf8]{inputenc}

\usepackage{microtype}

\usepackage{graphicx}
\usepackage{graphics}
\usepackage{hyperref}
\usepackage{url}
\usepackage{multirow}

\usepackage{colortbl}
\usepackage{makecell}

\usepackage{booktabs}
\usepackage{arydshln}
\usepackage{tabularx}
\newcolumntype{C}{>{\centering\arraybackslash}X}
\usepackage{footmisc}
\usepackage{lipsum}

\definecolor{TableRed}{HTML}{800000}
\definecolor{mediumelectricblue}{rgb}{0.01, 0.31, 0.59}

\usepackage{mdwlist}
\usepackage[list=true]{subcaption}
\usepackage{amsmath,amsfonts,amssymb}
\usepackage{pifont}

\usepackage{xspace}

\newcommand{\method}{\textsc{DeepStruct}\xspace}
\newcommand{\texthrt}[1]{\textsl{#1}}
\newcommand{\textspt}[1]{\texttt{#1}}
\newcommand{\comm}[1]{}

\newif\ifshowcomment
\showcommentfalse
\ifshowcomment
    \newcommand{\todo}[1]{\textcolor{yellow}{[TODO: #1]}}
    \newcommand{\dawn}[1]{\textcolor{purple}{[{Dawn: #1}]}}
    \newcommand{\jie}[1]{\textcolor{magenta}{[{Jie: #1}]}}
    \newcommand{\xiao}[1]{\textcolor{green}{[Xiao: #1]}}
    \newcommand{\zui}[1]{\textcolor{red}{[Zui: #1]}}
    \newcommand{\haoyun}[1]{\textcolor{blue}{[Haoyun: #1]}}
    \newcommand{\focus}[1]{{}}
\else
    \newcommand{\todo}[1]{}
    \newcommand{\dawn}[1]{}
    \newcommand{\jie}[1]{}
    \newcommand{\xiao}[1]{}
    \newcommand{\zui}[1]{}
    \newcommand{\haoyun}[1]{}
    \newcommand{\focus}[1]{}
\fi

\title{\method: Pretraining of Language Models for Structure Prediction}

\author{Chenguang Wang$^{*\dagger}$, Xiao Liu$^{\P\dagger}$, Zui Chen$^{\P\dagger}$, Haoyun Hong$^\P$, Jie Tang$^\P$, Dawn Song$^*$ \\
$^*$UC Berkeley, $^\P$Tsinghua University \\
\texttt{\{chenguangwang,dawnsong\}@berkeley.edu}, \texttt{jietang@tsinghua.edu.cn} \\
\texttt{\{liuxiao21,chenzui19,honghy17\}@mails.tsinghua.edu.cn} \\
}

\begin{document}
\maketitle

\def\thefootnote{$^\dagger$}\footnotetext{Equal contribution.}
\def\thefootnote{\arabic{footnote}}

\begin{abstract}
    We introduce a method for improving the structural understanding abilities of language models. Unlike previous approaches that finetune the models with task-specific augmentation, we pretrain language models on a collection of task-agnostic corpora to generate structures from text. Our structure pretraining enables zero-shot transfer of the learned knowledge that models have about the structure tasks. We study the performance of this approach on 28 datasets, spanning 10 structure prediction tasks including open information extraction, joint entity and relation extraction, named entity recognition, relation classification, semantic role labeling, event extraction, coreference resolution, factual probe, intent detection, and dialogue state tracking. We further enhance the pretraining with the task-specific training sets. We show that a 10B parameter language model transfers non-trivially to most tasks and obtains state-of-the-art performance on 21 of 28 datasets that we evaluate.\footnote{\label{ft:opensource}The code and datasets are available at \url{https://github.com/cgraywang/deepstruct}.}
\end{abstract}
\section{Introduction}
\begin{figure}[h]
    \centering
    \includegraphics[width=0.45\textwidth]{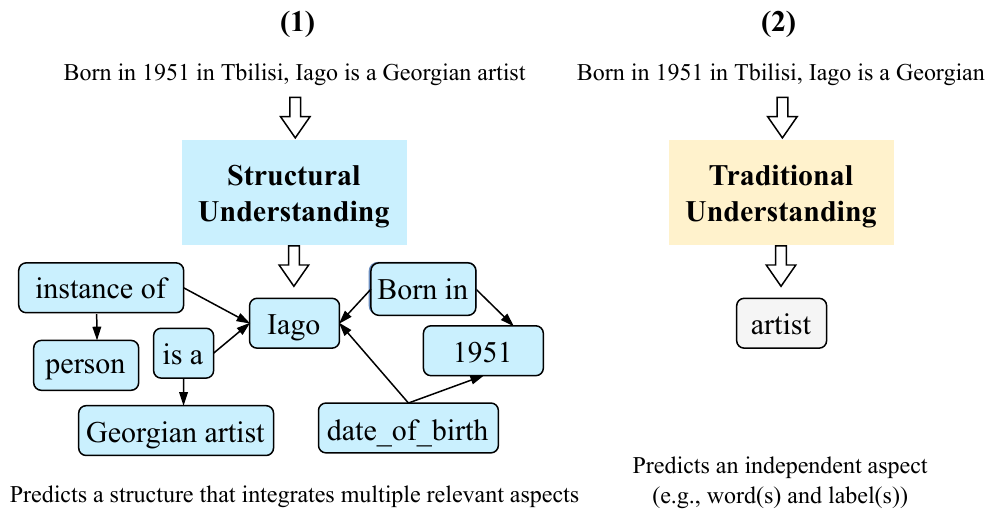}
    \caption{{\small {Comparison between structural understanding and traditional understanding of text.}}
      \label{fig:problemcomp}}
\end{figure}

Pretrained language models (LMs) have revolutionized NLP over the last few years~\cite{peters2018deep,devlin2019bert,radford2019language}, increasingly adept in performing the flexible and task-agnostic downstream transfer. Their transfer performance is less studied in structure prediction tasks, however. Well-studied tasks mainly focus on understanding one particular aspect of the text, such as predicting the next word that comes after as in language modeling. Unlike those downstream tasks, structure prediction requires the structural understanding of the text for further integrating multiple relevant aspects into a structure. For instance, a typical structure prediction task, called open information extraction, seeks the entire structural information in a sentence (Figure~\ref{fig:problemcomp}). Different from traditional NLP tasks, structure prediction takes one step further and serves as a natural testbed for the structural understanding competence of LMs.
\begin{figure}[!t]
    \centering
    \includegraphics[width=0.45\textwidth]{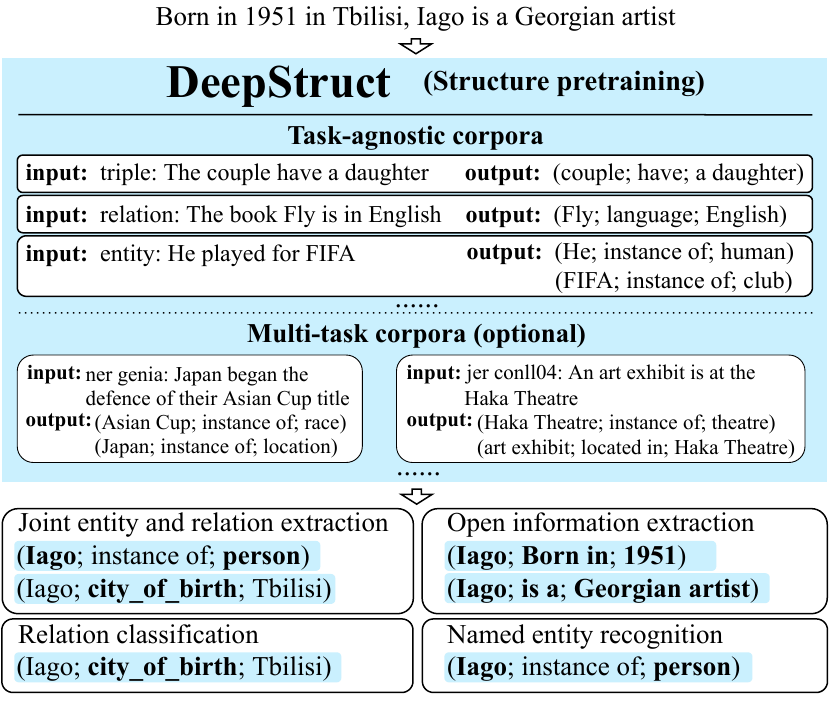}
    \includegraphics[width=0.48\textwidth]{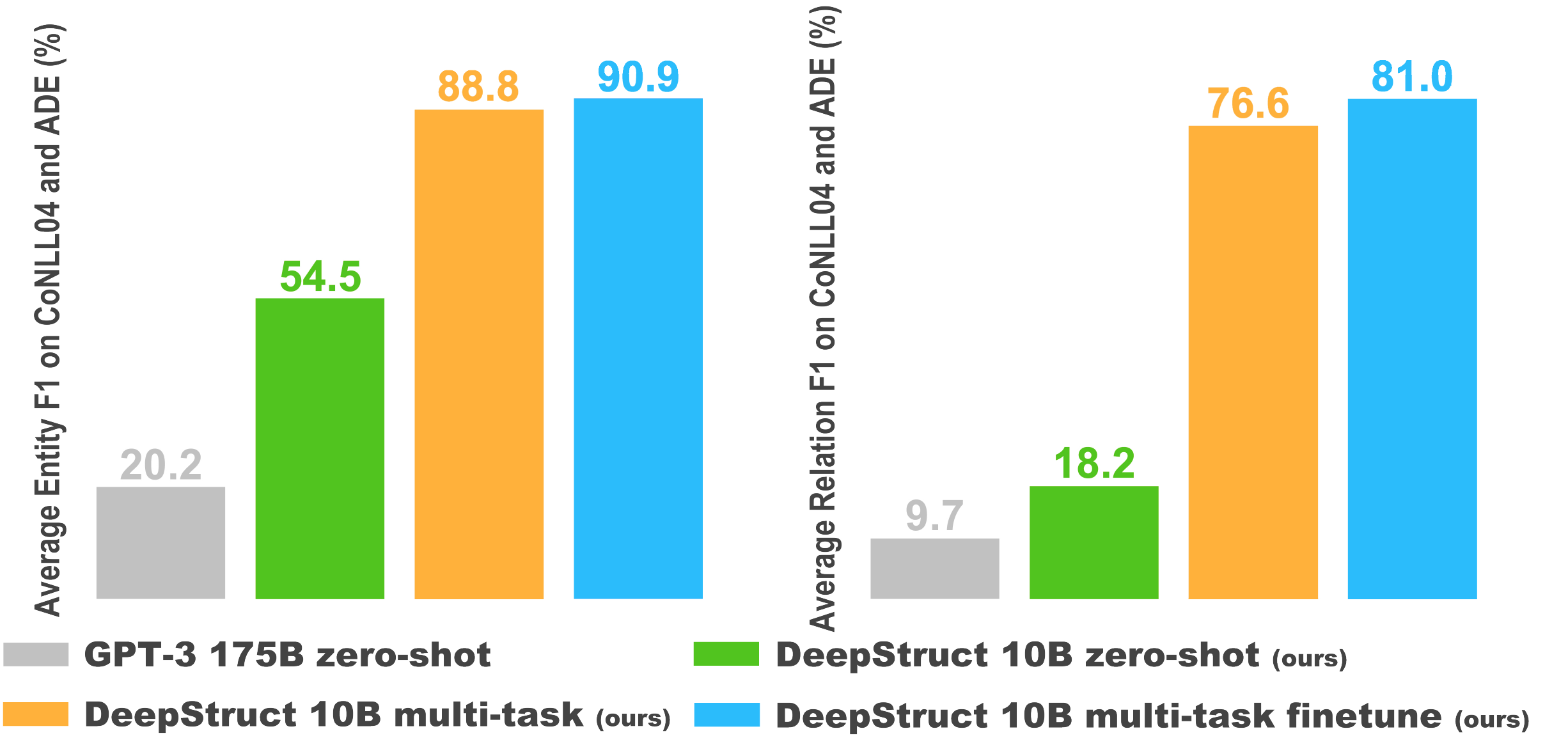}
    \caption{{\small {Summary of our approach and results. Upper: an overview of \method and the proposed structure pretraining. Lower: performance of our 10B \method\ zero-shot and multi-task, compared with 175B GPT-3 zero-shot.}}
      \label{fig:overview}}
    \vspace{-5mm}
\end{figure}

It is non-trivial to transfer LMs to downstream structure prediction tasks. While the structure prediction requires structural understanding, the LMs are pretrained to understand an independent aspect. For example, GPT-3~\cite{brown2020language} is trained to predict the next word, and BERT~\cite{devlin2019bert} is trained to recover the masked tokens. Recent work has made efforts in bridging the gap in transferring pretrained models to structure prediction tasks with a focus on two directions. As shown in Figure~\ref{fig:approach}, first, task-specific architectures are proposed to model the structures for different structure prediction tasks~\cite{stanovsky2018supervised,soares2019matching}. Second, task-specific data augmentation~\cite{paolini2021structured,wang2021zero,wei2021finetuned} is introduced, aiming to enrich text format with structure information. These approaches involve custom-designed task augmentations, impeding their usability in general structure prediction tasks.

In this paper, we improve the structural understanding capabilities of LMs. In contrast to previous approaches relying on task augmentations, we introduce structure pretraining, which systematically teaches LMs to better understand structures of text beyond independent aspects in a pretraining phase (Figure~\ref{fig:overview}). This enables the zero-shot transfer of knowledge that LMs learned about structures during our pretraining to downstream structure prediction tasks. For example, our zero-shot 10B parameter LM significantly outperforms the zero-shot GPT-3 (175B) on a structure prediction benchmark dataset (Figure~\ref{fig:overview}). We accomplish this by reformulating structure prediction as a series of unit tasks--triple prediction tasks. We then train LMs on a collection of task-agnostic structural corpora to generate triples from text. The design of triple representation is important: it unifies a wide set of standard structure prediction tasks into the same task format. We apply our pretrained model \method\ to 28 datasets spanning 10 structure prediction tasks, including open information extraction, joint entity and relation extraction, named entity recognition, relation classification, semantic role labeling, event extraction, coreference resolution, factual probe, intent detection, and dialogue state tracking. We further enhance the pretraining with multiple downstream structure prediction training sets and obtain state-of-the-art performance on 21 of 28 datasets. Our contributions are as follows:

\begin{itemize}[leftmargin=*]
    \item We improve structural understanding abilities of pretrained LMs. Compared to traditional NLP tasks that only consider the understanding of an independent aspect of the text, structural understanding takes a step further that requires the ability to integrate multiple relevant aspects into a structure. We argue that it is important for LMs to go beyond traditional understanding toward structural understanding, as it requires a higher level of intelligent competence and is more challenging. It can also benefit a wide spectrum of NLP tasks that require structure-level understanding capability.
    \item We propose structure pretraining, which pretrains the LMs to understand structures in the text. The basic intuition is that the standard pretraining helps LMs to understand individual aspects of the information in the text, our method learns to integrate those individual aspects into structures. Compared to existing approaches, this method enables the zero-shot transfer of LMs to structure prediction tasks. For instance, our 10B LM produces superior zero-shot performance compared to 175B GPT-3 on a representative structure prediction task. 
    \item We further equip our pretraining with multi-task learning and apply our method to 28 structure prediction datasets across 10 tasks. We achieve state-of-the-art performance on 21 of 28 datasets that we evaluate. We hope this can help facilitate the structural understanding research in the NLP community.
\end{itemize}

\section{Structure Pretraining}

\begin{figure}[ht]
    \centering
    \includegraphics[width=0.48\textwidth]{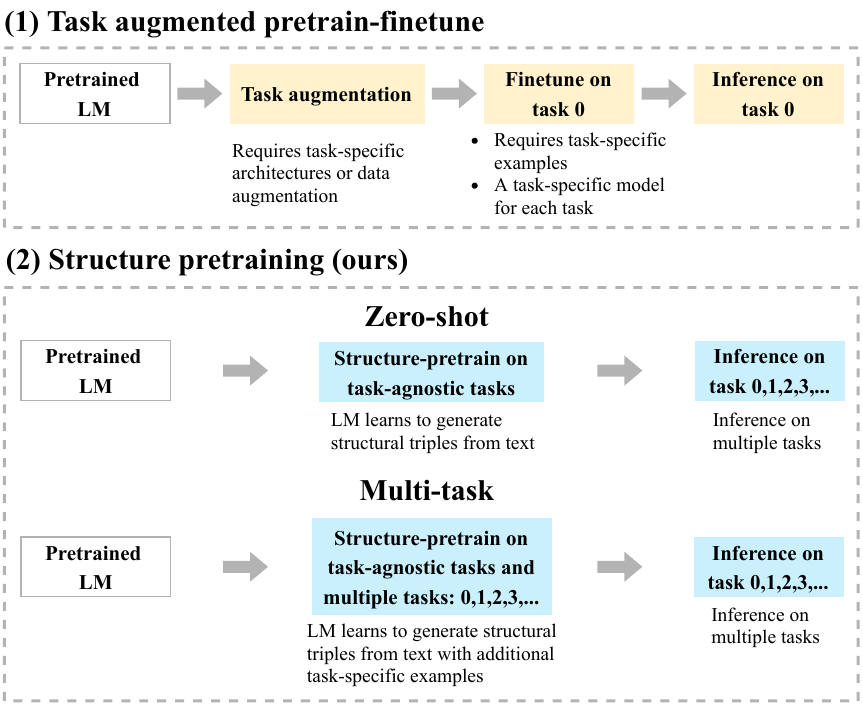}
    \caption{{\small {Comparing structure pretraining with standard pretrain-finetune paradigm.}}
      \label{fig:approach}}
\end{figure}

\label{sec:structpretrain}
The goal of our method is to improve the structural understanding capabilities of language models (LMs), i.e., understanding the structures of text. As shown in Figure~\ref{fig:approach}, instead of using the standard pretrain-finetune paradigm for each task, we introduce structure pretraining that aims to teach LMs to correspond to structures in a wide spectrum of tasks at the same time. We evaluate the structural understanding ability on multiple structure prediction tasks.

\subsection{Generative Pretraining} \label{sec:genp}
While the LM is pretrained to understand a single aspect of the text, structural understanding aims to recover the entire structure of the text (Figure~\ref{fig:problemcomp}). Structure pretraining is designed to bridge the gap via guiding LMs to produce structures from the text. It is ideal to generate arbitrary structures as needed. However, this is infeasible due to the highly complex nature of such structures. 

As an alternative, we reformulate the structure prediction as a combination of triple generation tasks. We refer to a triple as \texthrt{(head entity; relation; tail entity)} describing relations between entities. We design three pretraining tasks with a focus on predicting the entities, relations, and triples respectively. As shown in Figure~\ref{fig:overview}, (\expandafter{\romannumeral1}) entity prediction aims to output triples regarding the entities and their types in an input sentence. We implement this via prepending ``\textspt{entity:}'' as a prefix in the input. (\expandafter{\romannumeral2}) Relation prediction aims to recover the relations and corresponding types in the input as a triple. Similarly, we add ``\textspt{relation:}'' including a task separator ``\textspt{:}'' to each input. (\expandafter{\romannumeral3}) Triple prediction outputs the entire triple structure from the input. We attach ``\textspt{triple:}'' to indicate this task. These pretraining tasks are task-agnostic to downstream tasks, enabling the zero-shot downstream transfer (Sec.~\ref{sec:zs}).

Although the triple formulation is straightforward, we find that it is very flexible and able to model all structure prediction tasks we consider. A structure prediction task can be generally decomposed into generating the entities, relations, or triples. For example, named entity recognition predicts the entities and their types. It can be naturally represented as an entity prediction problem. Besides, traditional structure prediction tasks focusing on relations (e.g., relation classification) or triples (e.g., open information extraction) can be formulated as relation or triple prediction tasks respectively. A summary of all downstream tasks is described in Sec.~\ref{sec:task}.

We frame the pretraining as a conditional generation task where the input corresponds to text $x$, and the output $y$ is a sequence of triples. Our pretraining can be expressed as estimating a conditional distribution $p(y|x)$ in a probabilistic framework. We use an autoregressive LM to model $p(y|x)$.

\paragraph{Pretraining Data} We train the model on a collection of task-agnostic corpora including pre-built large-scale alignments between text and triples. In particular, we use T-REx~\cite{elsahar2018t}, TEKGEN and KELM~\cite{agarwal2020knowledge}, WebNLG~\cite{gardent2017webnlg}, ConceptNet~\cite{speer2012representing}. These corpora align text to triples consisting of high-quality entities and relations in knowledge graphs (e.g., Wikidata), which are used for entity and relation prediction tasks. In addition, for triple prediction tasks, we use OPIEC~\cite{gashteovski2019opiec} that provides open schema triples. The pretraining data statistics and the corresponding pretraining tasks are shown in Table~\ref{tab:ptdata}. Appendix~\ref{sec:implementation} shows additional details of our pretraining data.
\begin{table}[t]
\centering
\footnotesize
\renewcommand\tabcolsep{4.5pt}
\resizebox{0.98\linewidth}{!}{
\begin{tabular}{lllll}
\toprule[1.2pt]
\multirow{2}{*}{\textbf{Dataset}} & \multirow{2}{*}{\textbf{\#Sent.}} &  \multirow{2}{*}{\makecell[l]{\textbf{\#Rel.}\\(\#Tri.)}} & \multirow{2}{*}{\textbf{Task}}  \\ 
&&&& \\\midrule
T-REx~\cite{elsahar2018t}         & 6.2M                             & 11.1M                          & entity, relation \\
TEKGEN~\cite{agarwal2020knowledge}        & 7.9M                              & 16M                          & entity, relation \\
KELM~\cite{agarwal2020knowledge}          & 18M                             & 45M                          & entity, relation \\
WebNLG~\cite{gardent2017webnlg}        & 85K                                & 261K                           & relation         \\
ConceptNet~\cite{speer2012representing}   & 610K                               & 610K                           & relation         \\
OPIEC~\cite{gashteovski2019opiec}         & 26.8M                             & 104M                          & triple           \\ \bottomrule[1.2pt]
\end{tabular}
}
\caption{{Pretraining dataset statistics and corresponding pretraining tasks. \#Sent. and \#Rel. denote the number of sentences and relations respectively.}}
\label{tab:ptdata}
\end{table}

\begin{figure}[!t]
    \centering
    \includegraphics[width=0.48\textwidth]{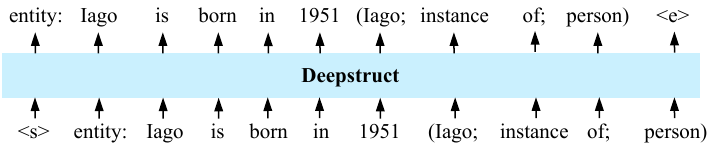}
    \caption{{\small {Summary of training procedure}.}
      \label{fig:train}}
\end{figure}
\begin{figure*}[!t]
    \centering
    \includegraphics[width=0.95\textwidth]{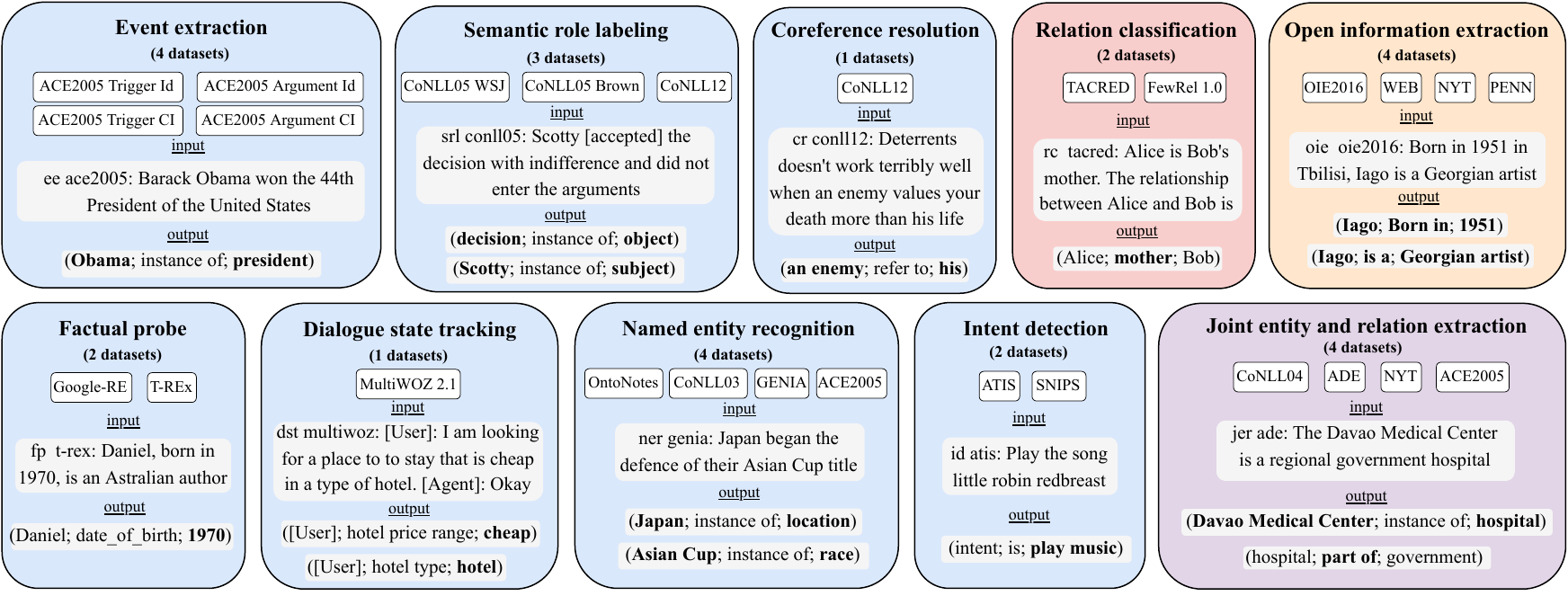}
    \caption{{\small {Summary of tasks and datasets. Blue: entity prediction task; Red: relation prediction task; Purple: entity and relation prediction task; Yellow: triple prediction task.}}
      \label{fig:tasks}}
    \vspace{-5mm}
\end{figure*}
Figure~\ref{fig:train} shows an example of the training procedure for the entity prediction task based on the input and output sample below.
\begin{enumerate*}
    \item[] {\bf Input} \textspt{entity}: Iago is born in 1951
    \item[] {\bf Output} ({\bf Iago}; instance of; {\bf person})
\end{enumerate*}
where the input text and output triple are aligned, and the alignment is provided by our pretraining data. Tokens are predicted autoregressively starting with \textspt{<s>} token and ending with \textspt{<e>} token. The head entity (i.e., Iago) and the tail entity (i.e., person) of the output triple then serve as the predictions (i.e., entity mention and entity type) of named entity recognition.

\subsection{Tasks}
\label{sec:task}
It is resource-intensive to create large-scale structural understanding datasets from scratch. Therefore, we collect existing datasets in the field of structure prediction for evaluation. We consider 28 datasets spanning 10 structure prediction tasks as shown in Figure~\ref{fig:tasks}. Detailed descriptions and sizes of datasets are shown in Appendix~\ref{appendix:exp}. 

\subsection{Zero-Shot}
\label{sec:zs}
The zero-shot \method\ refers to the setting where the pretrained model is used without any task-specific training at inference time. This differs from prior fully supervised methods. This setting is challenging as it might be difficult for humans to understand the tasks without prior examples. For example, if we are asked about ``semantic role labeling'' that aims to recover the predicate-argument structure, it is hard to understand what this really means. Nevertheless, the existing zero-shot setting resonates with human behaviors. For example, for named entity recognition, a human can understand and follow the instruction.

We enable the zero-shot transfer to structure prediction tasks via converting the downstream tasks to one or a combination of the pretraining tasks. As shown in Figure~\ref{fig:tasks}, at inference time, seven structure prediction tasks are formulated as entity prediction with the prefix ``\textspt{entity:}'' attached to the input example (in blue), while one task is cast as relation prediction with the prefix ``\textspt{relation:}'' (in red). In addition, open information extraction is a triple prediction task with the prefix ``\textspt{triple:}'' (in yellow). Joint entity and relation extraction (JER) is a combination of entity and relation prediction (in purple). For the entity and relation prediction, we use the prefix ``\textspt{entity:}'' and ``\textspt{relation:}'' respectively. Besides, for each dataset, we build a schema alignment between the pretraining dataset and downstream dataset (details are described in Sec.~\ref{sec:discussion}). The output triples are then decoded as corresponding structure predictions based on the pre-built schema alignment.

\subsection{Multi-Task} 
However, the distribution of the pretraining data is not perfectly aligned with the distribution of downstream datasets. This results in a shift in the output distribution of the pretrained model. The zero-shot setting cannot perform at its best on several out-of-distribution tasks including dialogue state tracking. The reason is that its desired output is a dialogue state, which is lacking in our task-agnostic pretraining corpora. To mitigate this, we integrate multiple structure prediction datasets into the pretraining corpora, and train our method on the mixture of the datasets. We list an example input and output format for each task in Figure~\ref{fig:tasks}. For all datasets of a particular task, we adopt the same input and output format. We also add task name and dataset name followed by the separator ``\textspt{:}'' as a prefix to each input example. For example, we add ``\textspt{jer ade:}'' to indicate one of the JER datasets, ADE. More examples of each task and dataset are shown in Table~\ref{tab:datasetexamples}. In contrast to fully pretrain-finetuned models that store a copy of parameters for each task, this setting is a lightweight alternative and produces a single model for all tasks, improving parameter sharing.

After multi-task training, we further finetune our method on the task-specific dataset for each task. The intuition is that finetuning is the de facto way to leverage pretrained LMs to perform downstream tasks. We aim to test an upper bound of the transfer performance of our structure pretraining via the additional finetuning phase. For both multi-task settings, we use the same data format with the training at test time. Basically, we add the task name and dataset name followed by the separator to the input example.

\section{Experiments}
In this section, we show that \method\ successfully transfers to the structure prediction tasks considered and obtain state-of-the-art results on 21 of 28 datasets we evaluate. All results are obtained via structure pretraining a pretrained 10B parameter LM, GLM~\cite{du2021all}. The details of the experimental setup, datasets, and comparison methods are described in Appendix~\ref{appendix:exp}.

\subsection{Main Results}

\begin{table*}[!t]
\footnotesize
\renewcommand\tabcolsep{4pt}
\renewcommand\arraystretch{0.92}
\resizebox{1.0\linewidth}{!}{
\begin{tabular}{@{}p{1.7cm}<{\centering}            p{2.2cm}        p{1.8cm}p{1.0cm}<{\centering}                                             p{0.25cm}<{\raggedright}        p{3.7cm}<{\raggedright}                                                 p{1.2cm}<{\centering}                 c                        c                               p{1.5cm}<{\centering}   @{}}
\toprule[1.2pt]
\multirow{3}{*}{Task}                             & \multicolumn{2}{l}{\multirow{3}{*}{Dataset}}& \multirow{3}{*}{Metric}                                   & \multicolumn{2}{l}{\multirow{3}{*}{\makecell[l]{Task-specific model}}}  & \multirow{3}{*}{TANL}             & \multicolumn{3}{c}{\bf \method}                                                       \\
\cmidrule(l){8-10}                                &                             &               &                                                                       &                               &                                                                       &                                   & zero-shot                 &  \multicolumn{2}{c}{multi-task}                           \\   
                                                  &                             &               &                                                                       &                               &                                                                       &                                   &                           &                               & w/ finetune               \\ \midrule
\multirow{4}{*}{\makecell[c]{Open\\information\\extraction}}
                                                  & OIE2016                     &               & \multirow{4}{*}{F1}                                                   &  67.0                         &~\cite{stanovsky2018supervised}                                        & -                                 & 28.1                      & \underline{71.2}              & \bf 71.3                  \\
                                                  & WEB                         &               &                                                                       & \bf 58.9                      &~\cite{stanovsky2016getting}                                           & -                                 & 43.8                      & \underline{50.8}              & 49.1                      \\
                                                  & NYT                         &               &                                                                       & 38.3                          &~\cite{saha2018open}                                                   & -                                 & 28.9                      & \underline{43.6}              & \bf 45.0                  \\
                                                  & PENN                        &               &                                                                       & 42.6                          &~(\textcolor{darkblue}{OpenIE4}~\footref{ft:openie51})                 & -                                 & \underline{51.0}          & \bf 54.5                      & 45.1                      \\ \midrule
\multirow{5}{*}{\makecell[c]{Relation\\classification}}
                                                  & TACRED                      &               & \multirow{5}{*}{F1}                                                   & 73.9                          &~\cite{sainz2021label}                                                 & 71.9                              & 36.1                      & \underline{74.9}              & \bf 76.8                  \\
                                                  & \multirow{4}{*}{FewRel 1.0} & 5-way 1-shot  &                                                                       & 90.1                          &~\cite{soares2019matching}                                             & 93.6$\pm$5.4                      & 72.4$\pm$6.9              & \underline{93.6$\pm$6.0}      & \bf 98.4$\pm$2.8          \\
                                                  &                             & 5-way 5-shot  &                                                                       & 89.5                          &~\cite{gao2019fewrel}                                                  & \underline{97.6$\pm$3.2}                      & 70.8$\pm$8.0              & 96.4$\pm$4.2      & \bf 100.0$\pm$0.0         \\
                                                  &                             & 10-way 1-shot &                                                                       & 83.4                          &~\cite{soares2019matching}                                             & 82.2$\pm$5.1                      & 67.6$\pm$4.5              & \underline{92.2$\pm$6.4}      & \bf 97.8$\pm$2.0          \\
                                                  &                             & 10-way 5-shot &                                                                       & 81.8                          &~\cite{gao2019fewrel}                                                  & 89.8$\pm$3.6                      & 66.4$\pm$6.3              & \underline{94.6$\pm$3.6}      & \bf 99.8$\pm$0.6          \\ \midrule
\multirow{8}{*}{\makecell[c]{Joint entity\\and relation\\extraction}}
                                                  & \multirow{2}{*}{CoNLL04}    &               & \multirow{8}{*}{\makecell[c]{F1\\($\frac{{\rm Ent.}}{{\rm Rel.}}$)}}  & 88.9                          &\multirow{2}{*}{~\cite{zhao-etal-2020-asking}}                         & \underline{90.3}                  & 48.3                      & 88.4                          & \bf 90.7                  \\
                                                  &                             &               &                                                                       & {71.9}              &                                                                       & 71.4                              & 25.8                      & \underline{72.8}                          & \bf 78.3                  \\
                                                  & \multirow{2}{*}{ADE}        &               &                                                                       & 89.3                          &\multirow{2}{*}{~\cite{DBLP:journals/corr/abs-1909-07755}}             & \bf 91.2                          & 60.7                      & 90.5                          & \underline{91.1}          \\
                                                  &                             &               &                                                                       & 78.8                          &                                                                       & \bf 83.8                          & 10.6                      & \underline{83.6}              & \bf 83.8                  \\
                                                  & \multirow{2}{*}{NYT}        &               &                                                                       & -                             &\multirow{2}{*}{~\cite{yuan2020relation}}                              & 94.9                              & 60.5                      & \underline{95.4}              & \bf 95.9                  \\
                                                  &                             &               &                                                                       & 84.6                          &                                                                       & 90.8                              & 28.6                      & \bf 93.7                      & \underline{93.3}          \\
                                                  & \multirow{2}{*}{ACE2005}    &               &                                                                       & 88.4                          &\multirow{2}{*}{~\cite{luan2019general}}                               & 88.9                  & 31.8                      & \bf 90.2                          & \underline{90.0}                  \\
                                                  &                             &               &                                                                       & 63.2                          &                                                                       & \underline{63.7}                  & 5.3                       & 58.9                          & \bf 66.8                  \\ \midrule
\multirow{4}{*}{\makecell[c]{Event\\extraction}}
                                                  & \multirow{4}{*}{ACE2005}    & Trigger Id    & \multirow{4}{*}{F1}                                                   & 72.5                          &~\cite{nguyen2019one}                                                  & \underline{72.9}                  & -                         & 72.7                          & \bf 73.5                  \\
                                                  &                             & Trigger Cl    &                                                                       & \bf 69.8                      &~\cite{nguyen2019one}                                                  & 68.5                  & -                         & \underline{69.2}                         & \bf 69.8                  \\
                                                  &                             & Argument Id   &                                                                       & \underline{59.9}                      &~\cite{nguyen2019one}                                                  & 50.1                              & -                         & \bf 67.5                          & 59.4          \\
                                                  &                             & Argument Cl   &                                                                       & 52.5                          &~\cite{wadden2019entity}                                               & 48.5                              & -                         & \bf 63.9             & \underline{56.2}                  \\ \midrule
\multirow{4}{*}{\makecell[c]{Coreference\\resolution\\}}
                                                  & \multirow{4}{*}{CoNLL12}    &               & MUC                                                                   & \bf 86.3                      &\multirow{4}{*}{~\cite{wu-etal-2020-corefqa}}                          & \underline{81.0}                  & -                         & 63.9                          & 74.9                      \\
                                                  &                             &               & B$^3$                                                                 & \bf 77.6                      &                                                                       & 69.0                              & -                         & 57.7                          & \underline{71.3}          \\
                                                  &                             &               & CEAF$_{\phi4}$                                                        & \bf 75.8                      &                                                                       & 68.4                              & -                         & 60.2                          & \underline{73.1}          \\
                                                  &                             &               & Ave. F1                                                               & \bf 79.9                      &                                                                       & 72.8                              & -                         & 60.6                          & \underline{73.1}          \\ \midrule
\multirow{2}{*}{\makecell[c]{Intent\\detection}}
                                                  & ATIS                        &               & \multirow{2}{*}{F1}                                                   & \bf 97.8                      &\multirow{2}{*}{~\cite{haihong2019novel}}                              & \underline{97.6}                              & -                         & 97.3                          & \bf 97.8                  \\
                                                  & SNIPS                       &               &                                                                       & \underline{97.4}              &                                                                       & \bf 98.7                          & -                         & \underline{97.4}                          & 97.3                      \\ \midrule
\multirow{3}{*}{\makecell[c]{Semantic\\role\\labeling}}
                                                  & CoNLL05 WSJ                 &               & \multirow{3}{*}{F1}                                                   & 88.8                          &\multirow{3}{*}{~\cite{shi2019simple}}                                 & 89.3                              & -                         & \bf 95.5                      & \underline{95.2}          \\
                                                  & CoNLL05 Brown               &               &                                                                       & 82.0                          &                                                                       & 84.1                              & -                         & \underline{92.0}              & \bf 92.1                  \\
                                                  & CoNLL12                     &               &                                                                       & 86.5                          &                                                                       & 87.7                              & -                         & \bf 97.2                      & \underline{96.0}          \\ \midrule
\multirow{4}{*}{\makecell[c]{Named\\entity\\recognition}}
                                                  & CoNLL03                     &               & \multirow{4}{*}{F1}                                                   & \bf 93.5                      &~\cite{yu2020named}                                                    & 91.7                              & 44.4                      & \underline{93.1}              & 93.0                      \\
                                                  & OntoNotes                   &               &                                                                       & \bf 90.4                      &~\cite{yan2021unified}                                                 & \underline{89.9}                  & 2.5                       & 87.6                          & 87.8                      \\
                                                  & GENIA                       &               &                                                                       & \underline{80.5}              &~\cite{yu2020named}                                                    & 76.4                              & 47.2                      & 80.2                          & \bf 80.8                  \\
                                                  & ACE2005                     &               &                                                                       & \bf 86.9                      &~\cite{li2020unified}                                                  & \underline{84.9}                  & 28.1                      & -                             & \bf 86.9                  \\ \midrule
\multirow{2}{*}{\makecell[c]{Dialogue state\\tracking}}
                                                  & \multirow{2}{*}{MultiWOZ 2.1}&              & Joint Acc.                                                            & \multirow{2}{*}{\bf 55.7}     &\multirow{2}{*}{~\cite{hosseiniasl2020simple}}                         & \multirow{2}{*}{51.4}             & \multirow{2}{*}{-}        & \multirow{2}{*}{53.5}         & \multirow{2}{*}{\underline{54.2}}\\ \midrule
\multirow{2}{*}{\makecell[c]{Factual probe}}      & Google-RE                   &               & \multirow{2}{*}{P@1}                                                  & 78.0                          &\multirow{2}{*}{~\cite{petroni2020context}}                            & -                                 & \bf 97.9                  & \underline{90.3}              & -                         \\
                                                  & T-REx                       &               &                                                                       & 62.6                          &                                                                       & -                                 & \bf 85.0                  & \underline{71.0}              & -                         \\ \bottomrule[1.2pt]
\end{tabular}
}
\caption{Results on all tasks. All evaluation scores are higher the better. TANL is introduced in \cite{paolini2021structured}. The \textbf{bold} denotes the best, and the \underline{underline} indicates the second best.}
\label{tab:allres}
\vspace{-0.1in}
\end{table*}

We have two settings as described in Sec.~\ref{sec:structpretrain}: zero-shot and multi-task. We also finetune the multi-task version on each downstream dataset. In total, we have three versions of \method. For comparison: we report the performance of TANL~\cite{paolini2021structured} when available. We also show the best performance among the task-specific models that are described in Appendix~\ref{appendix:exp}. 
Table~\ref{tab:allres} reports the results.

\begin{table}[!t]
\footnotesize
\renewcommand\tabcolsep{6.5pt}
\begin{tabular}{@{}lcccc@{}}
\toprule[1.2pt]
\multirow{2}{*}{Model} & \multicolumn{2}{c}{CoNLL04} & \multicolumn{2}{c}{ADE} \\ \cmidrule(l){2-3} \cmidrule(l){4-5}
                       & Ent.        & Rel.         & Ent.       & Rel.                \\ \midrule
GPT-3 175B zero-shot   & 34.7         & 18.1         & 5.8        & 1.3                  \\ \midrule
\qquad\qquad\qquad\ \ zero-shot      & 48.3         & 25.8         & 60.7       & 10.6                 \\ 
\textbf{\method}\ multi-task  & 88.4     & 72.8      & 90.5      & 83.6      \\
\qquad\qquad\qquad\quad\hspace{0.02in}\ w/ finetune & 90.7    & 78.3      & 91.1      & 83.8      \\ \bottomrule[1.2pt] 
\end{tabular}
\caption{Compare \method\ to GPT-3~\cite{brown2020language} 175B zero-shot on CoNLL04 and ADE datasets (joint entity and relation extraction). Ent. and Rel. denote entity F1 and relation F1 respectively.}
\label{tab:zero-shot}
\end{table}

With the zero-shot setting, a single model is used to perform on multiple tasks without the need of any task-specific training. This is in contrast to previous approaches that rely on task-specific models and datasets for each task. In Table~\ref{tab:zero-shot}, we also report the zero-shot performance of GPT-3 175B~\cite{brown2020language} on CoNLL04 and ADE (JER) via formulating the task as a question answering problem through prompting (details of the formulation are described in Sec.~\ref{sec:discussion}). JER requires the model to extract a set of entities and a set of relations between pairs of entities from the input text. Each predicted entity or relation has to be also assigned to an entity or a relation type. Zero-shot \method\ 10B outperforms zero-shot GPT-3 175B by a large margin without any prompt engineering. As shown in Table~\ref{tab:allres}, overall, \method's zero-shot performance is still far from that of task-specific supervised models on most tasks. The only exception is that the zero-shot setting obtains the new state-of-the-art performance on the factual probe with averaging 20\% P@1 improvement. This is because the task-specific method is also zero-shot. Note that we have removed the overlapped sentences with the T-REx test sets (factual probe) from our pretraining data. The result indicates that the structure pretraining is able to adapt the LM knowledge to the tasks by making LM aware of symbolic knowledge in the pretraining corpora. Besides, the zero-shot approach generalizes well to all task-agnostic pretraining tasks including entity prediction (e.g., named entity recognition), relation prediction (e.g., relation classification), and triple prediction (e.g., open information extraction).

Similar to the zero-shot setup, we only train a single model to conduct all the downstream tasks under the multi-task setting. This is different from the supervised models that use task-specific models and parameters. We achieve state-of-the-art performance on three datasets. For the other datasets, we obtain a competitive performance within a few points of the best-compared methods. Notably, most structure prediction tasks show a large gain from zero-shot to multi-task. This suggests that most tasks are out-of-distribution of our structure pretrained model. Nevertheless, our method appears to be able to adapt to the downstream distributions, presenting a fair and strong performance with multi-task learning. Another explanation is that multi-task examples help the model better understand the downstream tasks, such as the output format of each task. We also observe strong multi-task performance on FewRel, which is a low-resource structure prediction benchmark. This suggests that the multi-task setting is beneficial in low-resource regimes via transferring knowledge from similar tasks. For our multi-task training, we leave out the ACE2005 named entity recognition dataset due to the overlap between train and test splits for different tasks. After finetuning, we obtain state-of-the-art performance on 21 datasets. For instance, we obtain a +8.0 absolute improvement and a +2.9 absolute improvement on CoNLL05 Brown (semantic role labeling) and TACRED (relation classification) compared to the state-of-the-art methods.

All above results are based on a pretrained 10B parameter LM, GLM. GLM is an autoregressive LM. In addition, the context $x$ is encoded by a bidirectional encoder. In principle, generative LMs, such as T5~\cite{raffel2020exploring}, BART~\cite{lewis2019bart} and GPT-3~\cite{brown2020language}, can also be used with the proposed structure pretraining for the structure prediction tasks as well. We leave this as one of the future investigations.

\subsection{Ablation Studies} 
\label{sec:ablation}
\paragraph{Pretraining Strategies}
As the key question of our work is to investigate how structure pretraining improves the structural understanding ability of LMs, we examine how different pretraining strategies impact the downstream performance. We evaluate the below settings on the CoNLL04 (JER). The first two settings examine the relative importance of the pretraining data: (\expandafter{\romannumeral1}) With example-proportional mixing: We follow \cite{raffel2020exploring} with a mixing rate maximum of 10K to balance the different sizes of datasets. All other components are kept the same with \method\ multi-task with finetuning.  (\expandafter{\romannumeral2}) With entity and relation augmentation: We add special tokens ``\textspt{[]}'' to indicate the positions of the entities and relations in a sentence. Additional details are shown in Appendix~\ref{appendix:jer}. (\expandafter{\romannumeral3}) No pretrain, finetune: We remove structure pretraining, and only finetune the LM on CoNLL04. (\expandafter{\romannumeral4}) Zero-shot: We only use the task-agnostic datasets and exclude the multi-task datasets in the pretraining. (\expandafter{\romannumeral5}) Multi-task: We use the multi-task model without finetuning. (\expandafter{\romannumeral4}) and (\expandafter{\romannumeral5}) are the same with the zero-shot and multi-task settings in Sec.~\ref{sec:structpretrain}. (\expandafter{\romannumeral6}) Finetune: The multiple downstream datasets are excluded in the structure pretraining, but the model is finetuned on CoNLL04.

Table~\ref{tab:ablationmain} shows the results. First, the distribution of pretraining data does not significantly shift from that of most tasks. This limits the impact of the balanced strategy (\expandafter{\romannumeral1}). The data augmentation (\expandafter{\romannumeral2}) does not bring additional benefits to the downstream performance. This confirms that the key to the success of structure prediction is our formulation that narrows down a complex structure to a set of triple prediction tasks. This allows the pretraining to capture the entities and relations that are important for tasks. Second, removing the structure pretraining (\expandafter{\romannumeral3}) provides the most direct ablation of how much structure pretraining helps. Structure pretraining significantly improves the LM in structure prediction. This is due to the gap between LM pretraining and downstream structural understanding. For example, the distribution of structure prediction datasets is different from or is considered as out-of-distribution for the pretraining data. Structure pretraining improves the adaptation to those datasets. Next, similar to the findings in Table~\ref{tab:allres}, we find that both task-agnostic training sets (\expandafter{\romannumeral4}) and multi-tasks datasets (\expandafter{\romannumeral5}) contribute to the strength of structure pretraining. In particular, finetuning is still very important to improve the downstream performance~\cite{logan2021cutting}. However, it produces a task-specific model for each dataset instead of a unified model for all tasks as in our zero-shot or multi-task setup. Compared to only finetuning the model on a downstream dataset (\expandafter{\romannumeral6}), the multi-task setting obtains sizable improvements. This is because if the downstream dataset sizes are small such as of CoNLL04, multi-task learning can be extremely helpful in the low-resource regimes~\cite{paolini2021structured}. We conduct the above ablation studies using a base version of \method\ with 220M parameters.
\begin{table}[t]
\centering
\footnotesize
\begin{tabular}{@{}lll@{}}
\toprule
{\bf Method}        & {\bf Ent.} & {\bf Rel.} \\ \midrule
\method\ 220M multi-task finetune           & 90.7 &  75.7 \\
\hspace{0.1in} With example-proportional mixing    & 88.0 & 73.1 \\
\hspace{0.1in} With entity and relation augmentation        & 88.6 & 74.9  \\
No pretrain 220M, finetune & 84.7 & 63.5 \\
\method\ 220M zero-shot           & 51.5 &  22.9   \\
\method\ 220M multi-task               & 76.9 & 55.2  \\
\method\ 220M finetune       & 87.4 & 70.4 \\
\bottomrule

\end{tabular}
    \vspace{-0.1in}
\caption{{\small Ablation over different facets of structure pretraining on CoNLL04 test set (joint entity and relation extraction). Ent. and Rel. indicate entity F1 and relation F1 respectively.}}
\label{tab:ablationmain}
    \vspace{-0.1in}
\end{table}

\begin{figure*}[ht]
    \centering
    \includegraphics[width=0.8\textwidth]{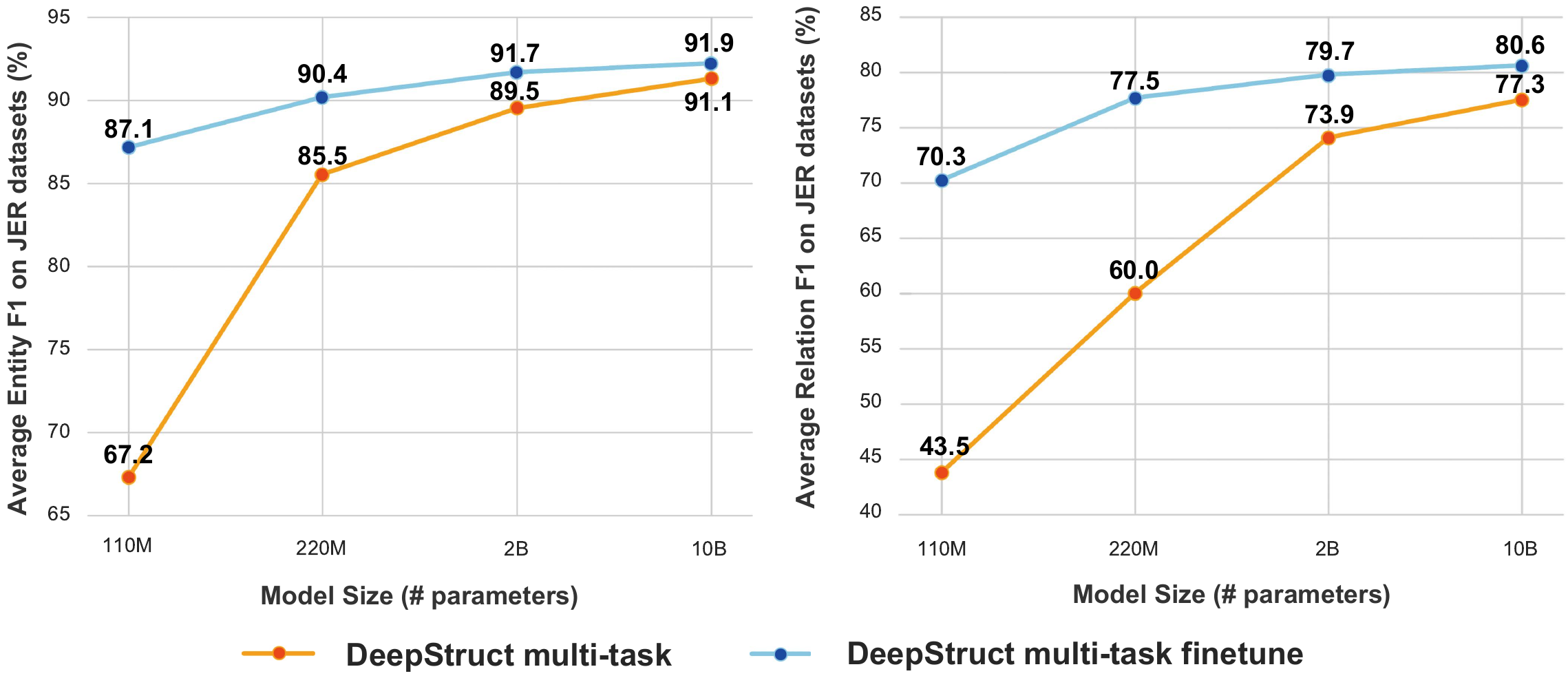}
    \vspace{-0.1in}
    \caption{{\small {Model scaling results on joint entity and relation extraction (JER) datasets. Left: entity F1; Right: relation F1.}}
      \label{fig:scaling}}
    \vspace{-0.1in}
\end{figure*}

\paragraph{Scaling Laws}
As it is often the case that larger models substantially improve the transferring capabilities of LMs~\cite{brown2020language,wei2021finetuned}, we explore how model scaling benefits the structure pretraining. We evaluate the effect on models with 110M, 220M, 2B, 10B parameters on JER datasets with multi-task and multi-task finetuned \method (Figure~\ref{fig:scaling}).

 As expected, average performance across the datasets improves as models grow larger. We find that when the models reach the order of 10B parameters, structure pretraining obtains the best performance. The 10B parameter model significantly improves the results compared to the 110M parameter model. One reason is that for small-scale models, learning across 28 structure prediction datasets during the structure pretraining may exceed the model capacity. For larger models, structure pretraining fully utilizes the model capacity and also teaches the models to generate triples according to the downstream tasks, allowing them to generalize well to most tasks with the rest capacity. It is also interesting that the performance does not seem to significantly saturate, indicating that the performance may further improve with larger-scale models. Under both setups, we observe similar trends. We also see that the model size matters more to the multi-task setting than to the finetuned version, suggesting finetuning is able to help specifically adapt to a task given a limited model size. The main pitfall is its generalization to more tasks.

\section{{Related Work}}
Pretrained LMs~\cite{devlin2019bert,radford2019language,yang2019xlnet} are the key ingredients in contemporary NLP. Sequence-to-sequence (seq2seq) LMs target conditional generation, such as T5~\cite{raffel2020exploring}, BART~\cite{lewis2019bart} and GLM~\cite{du2021all}. These models have benefited a wide range of nature language generation tasks such as summarization~\cite{zhang2020pegasus} and text infilling~\cite{zhu2019text,shen2020blank}. Recent attempts of generative prediction~\cite{paolini2021structured,schick2021s,lester2021power} have found that seq2seq models are able to provide a unified solution for modeling a wide set of NLP tasks. While existing approaches focus on text-to-text generation, \method\ aims to perform text-to-triple generation. 

Multi-task learning~\cite{caruana1997multitask} aims to train a model for multiple tasks simultaneously. In deep learning, it is usually categorized into hard weight sharing and soft weight constraints~\cite{ruder2017overview}. In the context of NLP, weight sharing has been adopted in \cite{collobert2008unified,yang2016multi,liu2019multi}. Since the emergence of large pretrained LMs~\cite{radford2018improving,devlin2019bert,yang2019xlnet}, multi-task training has been shown effective to enhance LMs' transferability to downstream tasks~\cite{raffel2020exploring}. Recent studies~\cite{wei2021finetuned} also show that pretrained models finetuned with abundant downstream tasks can conduct effective zero-shot learning. The main difference is that \method\ trains across multiple structure prediction datasets in structure pretraining with task-agnostic corpora, where we cast all datasets into triple formats.

Structure prediction is a long-standing challenge that relates to many NLP applications such as open information extraction~\cite{gashteovski2019opiec}, named entity recognition~\cite{tjong-kim-sang-de-meulder-2003-introduction,weischedel2013ontonotes}, and relation classification~\cite{zhang2017tacred,han2018fewrel,gao2019fewrel}. To handle different structure prediction problems, prior work presents a variety of task-specific models in the form of sequence tagging~\cite{stanovsky2018supervised,li2019dependency}, machine reading comprehension~\cite{zhao-etal-2020-asking} and text classification~\cite{soares2019matching}, which hinders the knowledge transfer across different tasks. TANL~\cite{paolini2021structured} proposes a translation-based approach to unify different structure prediction tasks with task-specific data augmentation. By contrast, our \method\ unifies more structure prediction tasks via a single model and a uniform data format.
\section{Discussion} \label{sec:discussion}
\paragraph{Related Models}
Recent studies have provided unified solutions for structural prediction tasks. We focus on the comparison between our \method\ to the state-of-the-art TANL~\cite{paolini2021structured} and DeepEx~\cite{wang2021zero}. TANL~\cite{paolini2021structured} proposes task-specific data augmentation (i.e., augmented natural language) that annotates task information and predictions in the input and output respectively for each structure prediction task. The main difference is that \method\ decomposes the structure prediction tasks into a collection of triple generation tasks. The triple format serves as the unified representation for all considered structure prediction tasks without the need of introducing new data augmentation as in TANL. While TANL mainly works in the multi-task setting, we additionally enable the zero-shot transfer via the task-agnostic structure pretraining. 
DeepEx~\cite{wang2021zero} explores the attention matrices of pretrained LMs via beam search to generate triples for information extraction tasks. Following the search, DeepEx introduces an extra ranking stage to improve the quality of the triples. Differently, \method\ aims to generate the triples for a wide set structure prediction tasks in an end-to-end fashion thanks to the proposed structure pretraining.

Besides, both TANL and DeepEx explore relatively small-scale pretrained LMs. Instead, \method\ scales up to 10 billion parameters. Figure~\ref{fig:scaling} shows that the performance improvements follow the scaling law~\cite{raffel2020exploring,lester2021power,wei2021finetuned,sanh2021multitask,liu2021p}. Based on our results, \method\ generalizes better to more structure prediction tasks compared to TANL and DeepEx.

\paragraph{Zero-Shot Setup}
For our zero-shot setup, we follow the zero-shot usage in recent pretrained LM studies~\cite{brown2020language,wei2021finetuned,sanh2021multitask}. It refers to the setting where a pretrained model is used to directly perform downstream tasks without including downstream training sets in its own pretraining data. For \method, our pretraining data is task-agnostic. For each task, we build an offline alignment between the schema of the pretraining data and the task dataset based on co-occurrence information in the pretraining data and downstream data~\cite{angeli2015leveraging}. We then manually curate the alignment. The resulting schema alignment is part of our release~\footref{ft:opensource}. At test time, we convert each task to one or a combination of the pretraining tasks based on Figure~\ref{fig:tasks}: entity, relation, or triple prediction. After producing the triples, we use the pre-built schema alignment to obtain the task predictions.

\begin{figure}
    \centering
    \includegraphics[width=0.45\textwidth]{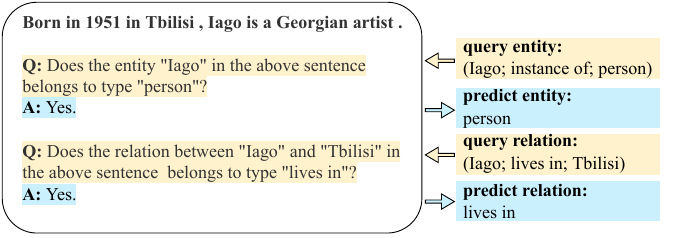}
    \vspace{-0.1in}
    \caption{{\small An example of GPT-3 zero-shot setting. To predict entities, we convert the gold entity triple (Iago; instance of; person) to an entity based true-or-false question. Similarly, to predict relations, the gold relation triple (Iago; lives in; Tbilisi) is turned into a relation based true-or-false question. The task predictions are correct if the answers are ``yes''.}}
    \vspace{-0.1in}
    \label{fig:gpt3}
\end{figure}

For GPT-3 zero-shot setting, we follow the prompting method introduced by GPT-3~\cite{brown2020language}. In more details, we aim to test the upper bound performance of GPT-3 for structure prediction, in particular the JER task. Therefore, instead of using standard prompts in the form of question answering, we design the prompts for ``true-or-false'' questions based on the ground truth. In this case, GPT-3 only answers with ``yes'' or ``no'' to produce a task prediction (Figure~\ref{fig:gpt3}), which is apparently an easier task compared to generating the structures from scratch.

\section{Conclusion}
We improve structural understanding capabilities of language models. We evaluate it on a wide set of structure prediction tasks including 10 tasks and 28 datasets, and successfully transfer pretrained language models to them through the proposed structure pretraining, which teaches language models to output triples from the text. We enable both zero-shot and multi-task transfer learning. \method\ obtains state-of-the-art results on 21 of 28 datasets. The result shows that pretrained language models can handle higher-level understanding (e.g., structural understanding), which may benefit more NLP tasks. We hope it will foster future research along the language structural understanding direction.

\section{Ethical Considerations}
We hereby acknowledge that all of the co-authors of this work are aware of the provided \textit{ACM Code of Ethics} and honor the code of conduct. 
This work is mainly about the pretraining and multi-task learning of LMs for structural prediction. 
The followings give the aspects of both our ethical considerations and our potential impacts to the community.
This work uses LMs, for which the risks and potential harms are discussed in ~\cite{brown2020language}.
There are potential undesirable biases that existed in task-agnostic data (e.g., from Wikipedia) and multi-task downstream datasets (mostly created from news articles).
We do not anticipate the production of harmful outputs, especially towards vulnerable populations, after using our model or training NLP models on our datasets.

\section{Environmental Considerations}
We adopt the pretrained LMs from the~\cite{du2021all}, whose energy cost and carbon
footprint during pretraining were 80.6 MWh and 4.6 tCO2e, respectively. Additionally, the
structure pretraining takes less than 5\% gradient-steps of the number of pretraining
steps of LMs, and thus the estimated auxiliary cost for energy is comparatively smaller.
In addition, training and tuning pretrained LMs on a wide range of tasks and datasets consume a plenitude of energy and increase emissions of carbon dioxide. To alleviate the problem, in this work we make efforts to study the multi-task training, which only involves training on a combination of all datasets at once. Our results (e.g., Figure~\ref{fig:scaling}) show that, despite the gap between multi-task and multi-task finetune on smaller models, the performance gap becomes minor when the model size scales up to 10 billion parameters. This indicates that we can reduce energy consumption when training large pretrained models via employing multi-task training.
\section*{Acknowledgement}
We would like to thank the anonymous reviewers for their suggestions and comments. We would also like to thank Fankun Zeng and Zijie Liu for their helpful inputs. This material is in part based upon work supported by Berkeley DeepDrive and Berkeley Artificial Intelligence Research. Xiao Liu, Zui Chen, Haoyun Hong, and Jie Tang are supported by the NSFC for Distinguished Young Scholar (61825602) and NSFC (61836013).

\bibliography{custom}
\bibliographystyle{acl_natbib}

\appendix
\section{Experimental Setup}
\label{appendix:exp}
\subsection{Implementation Details} 
\label{sec:implementation}

\paragraph{Model Architecture}
We leverage the Generalized Language Model (GLM)~\cite{du2021all} as our base language model pretrained on autoregressive blank infilling objectives. GLM follows an adaptive encoder-decoder architecture. It improves the pretrain-finetune consistency via cloze-style finetuning. GLM adopts the Byte Pair Encoding~\cite{radford2019language}, covering 50,257 tokens. In this work, we leverage the models in four different scales: 110M, 220M, 2B, and 10B~\footnote{https://github.com/THUDM/GLM}. The 110M model is pretrained over English Wikipedia and BookCorpus, and the others are pretrained over the Pile corpora~\cite{pile}. The Pile corpora are regarded as the similar corpora for training GPT-3. GLM outperforms T5 on text summarization, which shares a similar nature with structure prediction tasks. Compared to GPT-3, GLM is a bidirectional model and is able to perform autoregressive generation.

\paragraph{Structure Pretraining Procedure} 
\paragraph{Task-Agnostic Pretraining} We conduct the pretraining on 8 NVIDIA DGX-A100 machines using an Adam optimizer with a 5e-6 learning rate and 0.1 weight decay. We train the model with batch size 4 per GPU for 3 epochs and use the checkpoint of the last iteration.

\paragraph{Multi-Task Training} We conduct the multi-task training on 8 NVIDIA DGX-A100 machines using an Adam optimizer with a 5e-6 learning rate and 0.1 weight decay. We train the model with batch size 4 per GPU for 6 epochs and use the checkpoint with the best performance on the corresponding validation set for each dataset.

\paragraph{Inference} During the inference, length penalty and minimum target length are the most important hyperparameters. Length penalty is a float between 0 and 1 to control the GLM's generation length. The larger the length penalty is, the longer the generation length is. In general, for entity prediction tasks (e.g., NER, SRL, event extraction), a larger length penalty is used. For entity and relation prediction or triple prediction tasks (e.g., JER and OIE), a smaller one is used. For other tasks that require a specific number of output triples (e.g., relation classification, intent detection, factual probe), we trim the generation results according to the requirements of different tasks. We show details of the task-specific hyperparameters in Appendix~\ref{sec:appendixoie} to Appendix~\ref{sec:appendixid}.

\paragraph{Pretraining Data}
We apply the task-agnostic pretraining data presented in Sec.~\ref{sec:genp} during structure pretraining. A small portion of T-REx~\cite{elsahar2018t} is used in the factual probe task~\cite{petroni2020context}. To avoid the test leakage, we (\expandafter{\romannumeral1}) sample a small portion of the T-REx as our pretraining data, and (\expandafter{\romannumeral2}) remove samples that appeared in the T-REx dataset of the factual probe task from the pretraining data. We integrate WebNLG 2.1 and 3.0 into a WebNLG dataset. For OPIEC, We use its OPIEC-clean version. Similar to T-REx, we also sample a portion of the OPIEC for our pretraining due to its large size.

The following sections introduce the dataset formats, comparison methods, and training details for all 10 structure prediction tasks. We show additional input and output examples on all datasets in Table~\ref{tab:datasetexamples}.

\begin{table}[t]
\centering
\resizebox{\linewidth}{!}
  {
  \begin{tabular} {l | l | r r r}
    \toprule
    \multirow{2}{*}{{\bf Task}} & \multirow{2}{*}{{\bf Dataset}} & \multicolumn{3}{l}{{\bf \#Sents}} \\
    & & Train & Dev & Test \\
    \hline
    \multirow{4}{*}{{\small \bf Open information extraction}}
                                    & OIE2016                           &  2,278  &    571 &     589 \\
                                    & WEB                               &       - &      - &     920 \\
                                    & NYT                               &       - &    300 &     149 \\
                                    & PENN                              &       - &      - &      51 \\
    \hline
    \multirow{2}{*}{{\small \bf Relation classification}}
                                    & TACRED                            & 	68,124  & 22,631 &   15,509 \\
                                    & FewRel 1.0                            & 56,000  &      1,120 &  --  \\
    \hline
    \multirow{4}{*}{{\small \bf Joint entity and relation extraction}}    
                                    & CoNLL04                           &     922 &    231 &    288 \\
                                    & ADE                               &   3,845 &    --  &    427 \\
                                    & NYT                               &  56,195 &  5,000 &  5,000 \\
                                    & ACE2005                           &   7,477 &  1,789 &  1,517 \\
    \hline
    \multirow{2}{*}{{\small \bf Event extraction}}
                                    & ACE2005 Trigger                     &  11,178 &    649 &    642 \\
                                    & ACE2005 Argument                     &   4,450 &    531 &    612 \\
    \hline
                    {\small \bf Coreference resolution}
                                    & CoNLL12                           &   3,991 &  2,359 &  2,421 \\
    \hline
    \multirow{2}{*}{{\small \bf Intent detection}}
                                    & ATIS                              &  4,478  &    500 &     893 \\
                                    & SNIPS                             & 13,084  &    700 &     700 \\
    \hline
    \multirow{4}{*}{{\small \bf Semantic role labeling}}
                                    & CoNLL05                           &  39,832 &  3,206 &    --  \\
                                    & CoNLL05 WSJ                       &  39,832 &  3,206 &  5,221 \\
                                    & CoNLL05 Brown                     &  39,832 &  3,206 &    779 \\
                                    & CoNLL12                           &  89,549 & 32,397 & 21,499 \\
    \hline
    \multirow{4}{*}{{\small \bf Named entity recognition}}
                                    & CoNLL03                           &  14,041 &  3,250 &  3,453 \\
                                    & OntoNotes                         &  59,924 &  8,528 &  8,262 \\
                                    & GENIA                             &  14,824 &  1,855 &  1,854 \\
                                    & ACE2005                           &   7,299 &    971 &  1,060 \\
    \hline
    {\small \bf Dialogue state tracking}                                & MultiWOZ 2.1                      & 62,367  &  7,371 &   7,368 \\
    \hline
    \multirow{2}{*}{{\small \bf Factual probe}}
                                    & Google-RE                         &      -  &      - &     552 \\
                                    & T-REx                             &      -  &      - &   3,403 \\
    \bottomrule
  \end{tabular}
  }
\caption{{\small Statistics of downstream datasets.}}
\label{tab:statistics}
\end{table}

\subsection{Open Information Extraction}
\label{sec:appendixoie}
For OIE, given a sentence, we are asked to extract triples consisting of arguments and predicates. An example of the input and output format is as follows.

\paragraph{Input and Output Format}
\begin{enumerate*}
    \item[] {\bf Input} Born in 1951 in Tbilisi, Iago is a Georgian artist.
    \item[] {\bf Output} (Iago; Born in; 1951) (Iago; is a; Georgian artist)
\end{enumerate*}

\noindent where we extract arguments (e.g., Iago) and their predicates (e.g., Born in) in the form of triples as outputs from the input sentence.

\paragraph{Datasets} We evaluate the performance of the compared approaches on OIE benchmark datasets including {\bf OIE2016}~\cite{stanovsky2016creating}, a dataset converted from QA-SRL~\cite{he2015question} based on Newswire and Wikipedia; three datasets transformed from news corpus, including {\bf NYT}~\cite{mesquita2013effectiveness}, {\bf WEB}~\cite{mesquita2013effectiveness}, {\bf PENN}~\cite{xu2013open}. The statistics of all datasets are shown in Table~\ref{tab:statistics}. 

\paragraph{Comparison Methods} We compare our method \method\ to the following OIE systems presented in ~\cite{stanovsky2018supervised}: (\expandafter{\romannumeral1}) ClausIE~\cite{del2013clausie}: which leverages linguistic and grammatical knowledge to split clauses in a sentence and identify their roles, (\expandafter{\romannumeral2}) OpenIE 4~\footnote{\label{ft:openie51}\tiny\url{https://github.com/dair-iitd/OpenIE-standalone}}: which integrates SRLIE~\cite{christensen2011analysis} and Relnoun~\cite{pal2016demonyms} systems, (\expandafter{\romannumeral3}) PropS~\cite{stanovsky2016getting}: which focuses on prepositional phrases structure in sentences for OIE, (\expandafter{\romannumeral4}) RnnOIE~\cite{stanovsky2018supervised}: a supervised recurrent neural network based approach that views the OIE task as a sequence tagging problem. We additionally compare to MAMA using BERT$_{\rm{LARGE}}$ from \cite{wang2020language}, which proposes to leverage knowledge stored in attention matrices for OIE.

\paragraph{Training Details}
We train our model on the OIE2016 training set for 5 epochs during multi-task finetuning. The per GPU batch size is 4. During inference, for OIE2016, we choose a length penalty of 0.8. WEB, NYT, and PENN only contain the test sets. For these datasets, we use a length penalty of 0.5 and trim the outputs to only contain the first triple. We adopt the preprocessed OIE datasets provided by \citet{stanovsky2018supervised}.

\paragraph{Additional Results}
A detailed comparison between \method and compared approaches is shown in Table~\ref{tab:oie}. On OIE2016, NYT, and PENN datasets, \method presents significant improvements compared to the OIE systems. While on WEB, PropS~\cite{stanovsky2016getting} outperforms our method. The reason is that the arguments of WEB are very short and concise (e.g., ``( google ; assimilates ; youtube )''), which aligns better with the phrase extraction paradigm of PropS. We also observe that finetuning hurts the performance on WEB and PENN. This is because we are only able to finetune \method on the OIE2016 training set and this can lead to overfitting.

\subsection{Relation Classification}
Given head and tail entities in the target sentence, we seek to identify the relation between them. An example is as below.

\paragraph{Input and Output Format}
\begin{enumerate*}
    \item[] {\bf Input} The 1976 Thomas Cup was the tenth edition of Thomas Cup, the world championship of men's international team badminton (its female counterpart is the Uber Cup). The relationship between Uber Cup and badminton is
    \item[] {\bf Output} (Uber Cup; sport; badminton)
\end{enumerate*}

\noindent where ``Uber Cup'' and ``badminton'' are the corresponding head and tail entities, and ``sport'' is a relation from a predefined category. In addition, we augment the input sentence with the task-specific suffix ``The relationship between [head entity] and [tail entity] is'' following \cite{paolini2021structured} as shown in the above example.

\paragraph{Datasets} We evaluate on FewRel~\cite{han2018fewrel} and TACRED~\cite{zhang2017tacred}. 
\begin{itemize}
    \item {\bf FewRel} is a few-shot N-way K-shot relation classification dataset for meta learning. For all 100 relations, train (64 relations), validation (16 relations), and test set (20 relations) are constructed accordingly. We report the results on the dev set.
    \item {\bf TACRED} is a large-scale benchmark including over 100K samples and 41 relation types. We select the checkpoint on the dev set, and report results on the test set.
\end{itemize}
We show the dataset statistics in Table~\ref{tab:statistics}. 
We use F1 to evaluate the results. We parse every relation type and the corresponding head and tail entities from every original sample and formulate it as the input and output format as shown in the above example.

\paragraph{Comparison Methods} The compared models are as follows: (\expandafter{\romannumeral1}) BERT-PAIR~\cite{gao2019fewrel} is a sentence-level pairwise model that optimizes the similarity between sentences with the same relation. (\expandafter{\romannumeral2}) BERT$_{\rm EM}$+Matching the Blanks (MTB)~\cite{soares2019matching} proposes continual pretraining for relations over a large-scale entity-linked corpus. (\expandafter{\romannumeral3}) TANL~\cite{paolini2021structured} is a sequence-to-sequence generation model using task-augmented natural languages.  

\paragraph{Training Details} 
We train our model on training sets of TACRED and FewRel for 20 epochs during multi-task finetuning respectively. The per GPU batch size is 4. As shown in the above input and output format, we use the prompt of ``The relationship between [head entity] and [tail entity] is'' to query the model to generate the relation. For the zero-shot setting, the model is also provided with the prefix ``( [head entity];'' to generate the relation and tail entity. The prediction is correct only if both the relation and tail entity are correct. The length penalty equals 0.5.

\paragraph{Additional Results}
We show the results in Table~\ref{tab:rc}. \method outperforms all supervised methods on both TACRED and FewRel. We find that our task-agnostic pretraining can significantly help improve relation classification. This is vital to few-shot settings (FewRel), where \method can achieve almost perfect F1 scores. We also notice that multi-task \method outperforms all compared approaches except for the 5-5 FewRel setting.

\subsection{Factual Probe}
Given an input sentence, and a gold head entity and relation, the task is to predict the missing tail entity in the following.
\paragraph{Input and Output Format}
\begin{enumerate*}
    \item[] {\bf Input} Daniel Bowen, born in 1970, is a Melbourne resident best known as the author of the blog, Diary of an Average Australian.
    \item[] {\bf Output} (Daniel Bowen; date\_of\_birth; 1970)
\end{enumerate*}

\noindent where ``(Daniel Bowen; date of birth; '' is provided in the output, and the model is asked to generate ``1970)''.

\paragraph{Datasets} We use the Google-RE dataset consisting of 3 relations (``place of birth'', ``place of death'', and ``date of birth''), and T-REx with 41 relations from the LAMA benchmark~\cite{petroni2019language}. The task is evaluated using mean precision at one (P@1).

\paragraph{Comparison Methods} We compare to the following approaches: (\expandafter{\romannumeral1}) LAMA~\cite{petroni2019language} uses only the head and relation to form the query without using the oracle context, and (\expandafter{\romannumeral2}) LAMA-Oracle~\cite{petroni2020context} takes (at most) five gold sentences as additional context in the query. Both methods are based on BERT$_{\rm{LARGE}}$ and the query is constructed based on natural language templates. For example, the Wikidata relation ``place\_of\_birth'' is aligned with a template ``was born in''.

\paragraph{Training Details} 
As the factual probe task is usually performed without training sets, we only report \method's results in the zero-shot and multi-task setting (without finetuning). We follow the task format of LAMA-Oracle~\cite{petroni2020context}, which appends the query to five oracle context sentences. We use the relation labels (e.g., ``place of birth'') rather than the templates (e.g., ``was born in'') as they align better with our task-agnostic pretraining. Note that we have removed the T-REx data in the LAMA benchmark from the pretraining data.

\paragraph{Additional Results}
Table~\ref{tab:fr} shows the results. \method significantly outperforms compared approaches, which is attributed to the larger model size and knowledge-intensive task-agnostic pretraining. Multi-task setting actually hurts the performance due to the difference between the relation schema of downstream datasets and task-agnostic pretraining datasets. 

\subsection{Joint Entity and Relation Extraction}
\label{appendix:jer}
The goal of the task is to extract entities and relations (with their type information) from a given sentence. We formulate the task as two unit tasks: the first task is entity prediction to generate the entities, while the second task is relation prediction to generate the relations. Our task formalization is different compared to traditional JER, where our two unit tasks are independent. An example is as follows.
\paragraph{Input and Output Format}
\begin{enumerate*}
    \item[] {\bf Input} Blackstone already holds a 50 percent stake in the two parks that make up Universal Orlando.
    \item[] {\bf Entity Output} (Blackstone; instance of; organization) (parks; instance of; organization) (Universal Orlando; instance of; organization)
    \item[] {\bf Relation Output} (Blackstone; employer; parks)
\end{enumerate*}

\noindent where the entity output contains entity predictions and relation output contains relation predictions. The entity mentions are detected following the procedure below.

\paragraph{Evaluation Details} 
The conventional entity prediction evaluation is based on extractive span matching. To ensure a fair comparison in situations where there are multiple entities with the same surface, we adopt the following strategy: we match the spans of the generated entities from left to right in the original sentence when they are first mentioned. If there are duplicated entities, they are matched sequentially. For example, the first generated one matches the first mention span, while the second one matches the second mention span, etc. This strategy applies to all tasks that involve entity mention detection such as named entity recognition.

\paragraph{Datasets}
We experiment on the following datasets:

\begin{itemize}
    \item The {\bf CoNLL04}~\citep{roth-yih-2004-linear} dataset: CoNLL04 consists of four types of entities (``location'', ``organization'', ``person'', ``other'') and five types of relations (``work for'', ``kill'', ``organization based in``, ``live in'', ``located in'') on sentences taken from WSJ, AP, etc, containing 922 samples for training, 231 samples for validating, and 288 samples for testing. We use the same split as in~\cite{gupta-etal-2016-table}. We use the same type names of entities and relations with TANL~\cite{paolini2021structured}.
    \item The {\bf ADE}~\citep{GURULINGAPPA2012885} dataset: ADE contains annotated documents for drug-related adverse effects over medical case reports corpus. It consists of two entity types (``Adverse-Effect'' and ``Drug'') and one relation type (``(Has-)Adverse-Effect''). We use the same type names as in \cite{paolini2021structured}.
    \item The {\bf NYT}~\citep{10.1007/978-3-642-15939-8_10} dataset: NYT is a distantly-supervised joint entity and relation extraction dataset based on New York Times corpus. The dataset consists of three entity types (``PER'', ``ORG'', and ``LOC'') and 24 Freebase relations. We use a preprocessed version of this dataset from~\cite{yu2020jointer} and use the same type names with TANL~\cite{paolini2021structured}.
    \item The {\bf ACE2005}~\citep{walker2005ace} dataset: ACE2005 is based on the ACE 2005 Multilingual Training Corpus. We use a preprocessed version of this dataset in \cite{luan2019general}. We make use of seven entity types and six relation types with the same type names as in TANL.
\end{itemize}
The dataset statistics are shown in Table~\ref{tab:statistics}.

\paragraph{Comparison Methods}
We compare our method \method on the four datasets to the following JER methods: (\expandafter{\romannumeral1}) SpERT~\cite{DBLP:journals/corr/abs-1909-07755}: The BERT-based model first conducts named entity recognition formulated as sequence tagging, and performs relation classification between recognized entities; (\expandafter{\romannumeral2}) DyGIE~\cite{luan2019general}: The general information extraction framework organizes dynamic spans into graphs; (\expandafter{\romannumeral3}) MRC4ERE~\cite{zhao-etal-2020-asking}: The model formulates the joint entity and relation extraction task as machine reading comprehension; (\expandafter{\romannumeral4}) RSAN~\cite{yuan2020relation}: The work presents a relation-specific attention network to jointly extract entities and relations; (\expandafter{\romannumeral5}) TANL~\cite{paolini2021structured}: It is a sequence-to-sequence extraction model using augmented natural languages.

\paragraph{Training Details}
We train our model on JER training sets during multi-task finetuning for (\expandafter{\romannumeral1}) 10 epochs on CoNLL04, (\expandafter{\romannumeral2}) 10 epochs on ADE, (\expandafter{\romannumeral3}) 3 epochs on NYT, and (\expandafter{\romannumeral4}) 10 epochs on ACE2005. We employ less number of epochs on NYT, as its size is much larger compared to other datasets. We find that the relation prediction task and the entity prediction task need different length penalties. Therefore, we split the training sets corresponding to the two tasks. We choose a length penalty of 0.8 for entity prediction and 0.3 for relation prediction during inference. We use the same evaluation scripts as in \cite{paolini2021structured}. As multi-task training and finetuning for 10 folds on ADE is too expensive for \method 10B, only the first split of ADE is included in the multi-task training and finetuning.
In the ablation study (Sec.~\ref{sec:ablation}), we also present a model variant with entity and relation augmentation. For this setting, we augment the output with entity boundary information. For example, for the example in Figure~\ref{fig:overview}, ``([Iago]; instance of; person) ([Iago]; city\_of\_birth; [Tbilisi])'' are the augmented outputs. 

\paragraph{Additional Results}
Table~\ref{tab:jer} presents the results. \method outperforms or is competitive compared to state-of-the-art supervised approaches. We also find the multi-task \method performs competitively with previous task-specific approaches on both ADE and NYT. This indicates that multi-task trained models are cost-effective alternatives to per-task finetuned models.

\subsection{Named Entity Recognition}
Compared to joint entity and relation extraction, named entity recognition only focuses on predicting the entities and their corresponding types in the target sentence. We show an example below.

\paragraph{Input and Output Format}
\begin{enumerate*}
    \item[] {\bf Input} What we need to do is to make sure that state boards, number one, have adequate funding.
    \item[] {\bf Output} (we; instance of; human) (state; instance of; geographical entity) (state boards; instance of; organization)
\end{enumerate*}

\noindent where the head entities of these triples are the entity mentions in the given sentence, and tail entities are from a predefined list of entity types.

\paragraph{Datasets}
We experiment on the following datasets:
\begin{itemize}
    \item The {\bf CoNLL03}~\citep{tjong-kim-sang-de-meulder-2003-introduction} dataset: CoNLL03 (English) data was taken from the Reuters Corpus, containing 14,041 training samples, 3,250 validating samples and 3,453 testing samples. It consists four entity types (``LOC'', ``ORG'', ``PER'', and ``MISC''). We use the preprocessed version of this dataset from~\cite{li2020unified}.
    \item The {\bf OntoNotes}~\citep{pradhan-etal-2013-towards} dataset: OntoNotes contains 59,924 training samples, 8,528 validating samples, and 8,262 testing samples. It consists 18 different entity types (e.g., ``ORG'', ``PER''). We use the preprocessing scripts provided by \cite{luan2019general}.
    \item The {\bf GENIA}~\citep{10.5555/1289189.1289260} dataset: GENIA consists of compiled and annotated biomedical literature, which contains 14,824 training samples, 1,855 validating samples, and 1,854 testing samples. It consists five entity types (``DNA'', ``RNA'', ``cell\_line'', ``cell\_type'', and ``protein''). We use a preprocessed version of this dataset~\cite{li2020unified}.
    \item The {\bf ACE2005}~\citep{walker2005ace} dataset: ACE2005 contains 7,299 training samples, 971 validating samples, and 1,060 testing samples. Note that it is also processed based on the ACE2005 corpus but with different data splits compared to that of the ACE2005 JER dataset. It includes seven entity types. We use the preprocessed version of this dataset in \cite{li2020unified}, and exclude this dataset in the \method's multi-task setting due to its overlap with the ACE2005 JER dataset.
\end{itemize}

\paragraph{Comparison Methods}
We compare our method \method on the four datasets to the following NER methods: (\expandafter{\romannumeral1}) BERT-MRC~\cite{li2020unified}: this method formulates NER as a machine reading comprehension problem, (\expandafter{\romannumeral2}) BERT-MRC+DSC~\cite{li2020dice}: this model is a dice-loss enhanced version of BERT-MRC, (\expandafter{\romannumeral3}): Cloze-CNN~\cite{baevski2019cloze}: the model leverages cloze-style pretraining on convolutional neural networks for natural languages, (\expandafter{\romannumeral4}) GSL~\cite{athiwaratkun2020augmented}: the method uses augmented language for intent detection, slot filling, and named entity recognition, (\expandafter{\romannumeral5}) BiaffineLSTM~\cite{yu2020named}: the model transforms NER into dependency parsing using biaffine LSTMs, (\expandafter{\romannumeral6}) TANL~\cite{paolini2021structured}: it presents a sequence-to-sequence extraction approach using augmented natural languages.

\paragraph{Training Details} 
We train our model on NER training sets for 15 epochs on every dataset during multi-task finetuning with early stopping. The per GPU batch size is 4. We choose a length penalty of 0.8 during inference. Since some datasets may contain null predictions, we set the minimum target length to 0. 

\paragraph{Additional Results}
Table~\ref{tab:ner} shows the results. \method achieves comparable performance to task-specific supervised approaches, except for OntoNotes. We suppose that OntoNotes contains a relatively large number of entity types, making it more challenging for models to use labels for considering their semantic meanings. On GENIA and ACE2005, \method outperforms state-of-the-art task-specific methods.

\subsection{Semantic Role Labeling}
In semantic role labeling, we seek to identify the corresponding arguments in the form of spans (or the semantic roles) given a certain predicate. Consider an example as follow.

\paragraph{Input and Output Format}
The predicate is marked in the input. The model then yields arguments according to the predicate in the output with their corresponding argument types from a predefined set.
\begin{enumerate*}
    \item[] {\bf Input} Scotty [ accepted ] the decision with indifference and did not enter the arguments.
    \item[] {\bf Output} (Scotty; instance of; subject) (decision; instance of; object)
\end{enumerate*}

\noindent where ``[ accepted ]'' is the given predicate, and arguments such as ``Scotty'', ``the decision'' and their corresponding types are generated in the form of triples.

\paragraph{Datasets}
We experiment on the following datasets: CoNLL05 WSJ, CoNLL05 Brown \citep{carreras-marquez-2005-introduction} and CoNLL12 \citep{pradhan-etal-2013-towards}. Table~\ref{tab:statistics} shows the dataset statistics.

\begin{itemize}
    \item The {\bf CoNLL05 WSJ} and {\bf CoNLL05 Brown} datasets: CoNLL05 WSJ and CoNLL05 Brown datasets share the same train and validation splits. They have different test sets. For CoNLL05 WSJ and CoNLL05 Brown, the corresponding test datasets are taken from the WSJ and Brown corpus respectively. The datasets consist of seven different types including ``V'' (verb), ``A0'' (subject), ``A1'' (object), ``A2'', ``A3'', ``AM-MOD'', and ``AM-NEG''. We use the same type names as in ~\cite{paolini2021structured}. 
    \item The {\bf CoNLL12} dataset: CoNLL12 dataset is built upon OntoNotes dataset including 39 argument types. We leverage the same type names as in ~\cite{paolini2021structured}.
\end{itemize}

\paragraph{Comparison Methods}
We compare our method \method on the datasets to the following SRL models: (\expandafter{\romannumeral1}) Dep and Span~\cite{li2019dependency}: this model formulates semantic role labeling as an end-to-end dependency parsing task, (\expandafter{\romannumeral2}) BERT SRL~\cite{shi2019simple}: it is a sequence-tagging version of BERT, (\expandafter{\romannumeral3}) TANL~\cite{paolini2021structured}: this is a sequence-to-sequence extraction model using augmented natural languages. 

\paragraph{Training Details}
During multi-task finetuning, for CoNLL05, the model is trained on CoNLL05 WSJ's training set, and evaluated on both CoNLL05 WSJ and CoNLL05 Brown test sets. We train \method on CoNLL05 WSJ and CoNLL12 for 5 epochs respectively. The per GPU batch size equals 4. The length penalty is 0.8. 

\paragraph{Evaluation Details} Sentences with multiple target predicates are duplicated during data preprocessing. So, each sentence is only related to one target predicate that is marked by ``[]''. We adopt the same evaluation scripts as in ~\cite{paolini2021structured}.

\paragraph{Additional Results}
Table~\ref{tab:srl} shows the results. Both multi-task and multi-task finetuned \method outperform task-specific models by a large margin. An important reason is that PropBank~\cite{kingsbury2003propbank} is included in the multi-task training. The knowledge of PropBank transfers well to other SRL datasets. We find that the performance gain is significant since the large-scale model has the capacity to capture the PropBank knowledge. We also observe a minor performance drop from multi-task to multi-task finetuned \method on CoNLL05 WSJ and CoNLL12 datasets. This might be attributed to overfitting. Besides, the issue can be relieved if a better hyperparameter combination is used in the multi-task finetuning setting. 

\subsection{Event Extraction}
This task contains two sequential subtasks: (\expandafter{\romannumeral1}) event triggers identification and classification: this subtask first identifies the trigger words in target sentences that refer to certain types of events, and (\expandafter{\romannumeral2}) trigger arguments identification and classification: this subtask then extracts arguments from the target sentences that can be mapped to certain roles in the event from (\expandafter{\romannumeral1}).

\paragraph{Input and Output Format}
\begin{enumerate*}
    \item[] {\bf Trigger Input} But the Saint Petersburg summit ended without any formal declaration on Iraq .
    \item[] {\bf Trigger Output} (summit; instance of; meet)
    \item[] {\bf Argument Input} But the Saint Petersburg [ summit ] ended without any formal declaration on Iraq .
    \item[] {\bf Argument Output} (Saint Petersburg; instance of; place)
\end{enumerate*}

\noindent where ``summit'' is an extracted trigger and ``meet'' is its corresponding trigger event. Then, based on the ``summit'' event, we can further extract the role of ``place'' in this event as ``Saint Petersburg''.

\paragraph{Datasets}
We experiment on the ACE2005 dataset. For detailed dataset statistics, please refer to Table~\ref{tab:statistics}.
\begin{itemize}
    \item The {\bf ACE2005} dataset~\citep{walker2005ace}: ACE2005 contains 33 types of event triggers, and each of them corresponds to a set of argument roles. We follow the preprocessing in TANL~\cite{paolini2021structured} and use the same evaluation scripts. 
\end{itemize}

\paragraph{Comparison Methods}
We compare our method \method on the dataset to the following methods: (\expandafter{\romannumeral1}) J3EE~\cite{nguyen2019one}: This method presents a joint model based on a recurrent neural network to first extract mention spans for triggers and arguments and then perform pairwise classification, (\expandafter{\romannumeral2}) DyGIE++~\cite{wadden2019entity}: the method leverages BERT for sequence tagging to identify mention spans and then classify each mention with triggers in pair for argument roles, (\expandafter{\romannumeral3}) TANL~\cite{paolini2021structured}: this is a sequence-to-sequence extraction approach using augmented natural languages.

\paragraph{Training Details} 
We train our model on ACE2005 event trigger and argument training sets for 20 epochs during multi-task finetuning. The per GPU batch size is 4. During inference, we choose a length penalty of 0.8. The argument prediction task requires triggers as input to make predictions. An example is shown in Table~\ref{tab:datasetexamples}. For the argument prediction task, we first generate all trigger predictions using our 10B model. If there is more than one trigger in a sentence, we will duplicate the sentence to make sure that every sample corresponds to a single trigger.
The ACE2005 dataset is processed similarly to the named entity recognition task.

\paragraph{Additional Results}
Table~\ref{tab:ee} presents the results. \method is competitive with the state-of-the-art task-specific supervised models on trigger identification and classification, as well as the argument identification task. In the meantime, \method outperforms the comparison methods on the argument classification task. 

\subsection{Coreference Resolution}
The coreference resolution aims to identify and cluster mentions in a document that refers to the same entity. An example is as follows.

\paragraph{Input and Output Format}
\begin{enumerate*}
    \item[] {\bf Input} And deterrents don't work well when an enemy values your death more than his life.
    \item[] {\bf Output} (an enemy; refer to; his)
\end{enumerate*}

\noindent where ``an enemy'' appears as the target entity and ``his'' is the mention it refers to. ``refer to'' is provided as part of the output triple.

\paragraph{Datasets}
We experiment on the CoNLL12 \citep{pradhan-etal-2013-towards} dataset constructed from OntoNotes corpus. The dataset statistics are presented in Table~\ref{tab:statistics}. 

\paragraph{Comparison Methods}
We compare our method \method on the dataset to the following methods: (\expandafter{\romannumeral1}) Higher-order c2f-coref~\cite{lee2018higherorder}: this method proposes a fully differentiable formulation of coreference resolution via iterative refinement using attention mechanism, (\expandafter{\romannumeral2}) BERT+c2f-coref~\cite{joshi2019bert}: this model replaces original recurrent neural network backbone in ~\cite{lee2018higherorder} with BERT, (\expandafter{\romannumeral3}) CorefQA+SpanBERT~\cite{wu-etal-2020-corefqa}: the model formulates coreference resolution as question answering, (\expandafter{\romannumeral4}) TANL~\cite{paolini2021structured}: this is a sequence-to-sequence extraction model using augmented natural languages.

\paragraph{Training Details} 
We train our model on CoNLL12 coreference resolution training set for 40 epochs during multi-task finetuning. The per GPU batch size is 4. During inference, we choose a length penalty of 0.8. CoNLL12 coreference resolution has different evaluation metrics compared to other structure prediction tasks, including: (\expandafter{\romannumeral1}) MUC: a link-based metric that reflects the minimum number of missing mentions in the response chain~\cite{moosavi2016coreference}, (\expandafter{\romannumeral2}) B$^3$: a single-mention based metric which computes the macro F1 of all entity mentions, and (\expandafter{\romannumeral3}) CEAF$_{\phi 4}$: a similarity metric based on the assumption that the coreference map should be one-to-one.
Due to the limited maximum sequence length of language models, the dataset is chunked with a fixed size of 512 during data preprocessing. Following TANL~\cite{paolini2021structured}, only intra-chunk coreferences are preserved. We also use the same evaluation scripts with ~\cite{paolini2021structured}.

\paragraph{Additional Results}
Table~\ref{tab:coref} shows the results. \method presents better results compared to TANL and classic task-specific supervised approaches. However, \method fails when compared with the state-of-the-art coreference method. The main reason is that this task requires task-specific model architectures. In the meantime, we argue that it is promising to employ a unified framework for multiple structure prediction tasks.

\subsection{Dialogue State Tracking}
We are presented with a dialogue between a user and an agent to identify what information is known given a list of slots by the end of each round of the conversation. An example is as follows.

\paragraph{Input and Output Format}
\begin{enumerate*}
    \item[] {\bf Input} [User]: I would like a taxi from Saint Johns College to Pizza Hut Fen Ditton. [Agent]: What time do you want to leave and when you want to arrive? [User]: I want to leave after 17:15.
    \item[] {\bf Output} ([User]; taxi arrive by; not given) ([User]; taxi departure; Saint Johns College) ([User]; taxi destination; Pizza Hut Fen Ditton) ([User]; taxi leave at; 17:15)
\end{enumerate*}

\noindent in which by the end of this conversation, we know that the user wants to get to Pizza Hut Fen Ditton from Saint Johns College, leaving at 17:15, while the taxi's arrival time is unknown. The slots ``taxi arrive by'', ``taxi departure'', ``taxi destination'', and ``taxi leave at'' are provided for the output.

\paragraph{Datasets} We use the MultiWOZ 2.1 \cite{budzianowski2018large, ramadan2018large, eric2019multiwoz, zang2020multiwoz}, which is a daily dialogue dataset for task-oriented conversations. We follow the preprocessing in \citep{wu2019transferable}. Following TANL~\cite{paolini2021structured}, the ``police'' and ``hospital'' slots are excluded from the training set as they are absent in the test set. The resulting training set contains 7,904 samples. Dataset statistics are presented in Table~\ref{tab:statistics}.

\paragraph{Comparison Methods} We compare our method with: (\expandafter{\romannumeral1}) TRADE~\cite{wu2019transferable}: It is a transferable multi-domain generative dialogue state tracking model, (\expandafter{\romannumeral2}) SimpleTOD \cite{hosseiniasl2020simple}: This is a state-of-the-art task-specific model for dialogue state tracking based on GPT-2 \cite{radford2019language}. In addition, we also compare our method with TANL.

\paragraph{Training Details}
We finetune for 20 epochs. The maximum sequence length is 512, and the per GPU batch size is 4. Given a domain and all possible slots, \method generates triples regarding the slots: if the information is not yet provided, the tail should be ``not given''. We use the same type names with ~\cite{paolini2021structured}.

\paragraph{Additional Results}
Table~\ref{tab:dst} shows the results. While \method does not outperform the state-of-the-art SimpleTOD, it is still competitive compared to the task-specific supervised models. This demonstrates the effectiveness of \method in dealing with different structure prediction tasks under the same architecture.

\subsection{{Intent Detection}}
\label{sec:appendixid}
Intent detection identifies the user's intent in the conversation with the agent based on a predefined list of slots. It resonates with the classical sentence classification task. Below is an example.

\paragraph{Input and Output Format}
\begin{enumerate*}
    \item[] {\bf Input} Show flight and prices from Kansas City to Chicago next Wednesday arriving in Chicago by 7 pm.
    \item[] {\bf Output} (intent; is; flight and airfare)
\end{enumerate*}

\noindent where our prediction is the ``flight and airfare''. The head entity ``intent'' and predicate ``is'' are given for all outputs.

\paragraph{Datasets}
We use two datasets, the ATIS dataset \cite{hemphill1990atis}, which contains flight and airline-related conversation and queries, and the SNIPS dataset \cite{coucke2018snips}, which consists of daily queries from the interaction between the users and dialogue agents.
The dataset statistics are shown in Table~\ref{tab:statistics}.

\paragraph{Comparison Methods}
We compare our method to SF-ID~\cite{haihong2019novel} and TANL~\cite{paolini2021structured} in this task. 

\paragraph{Training Details}
We formulate the label of every sample as ``(intent; is; [label])''. We parse every intent from every original sample and formulate it into the input and output format as shown above. We finetune for 20 epochs. The maximum sequence length is 512, and the per GPU batch size is 4. We report F1 for this task.
            
\paragraph{Additional Results}
Table~\ref{tab:id} shows the results. \method is comparable to task-specific supervised approaches on both ATIS and SNIPS datasets. For TANL, we produce the results using the released code~\footnote{\tiny\url{https://github.com/amazon-research/tanl}}.

\begin{table*}[]
    \centering
    \small
    \renewcommand\tabcolsep{15.3pt}
    \begin{tabular}{@{}llllllll@{}} 
\toprule
\multicolumn{2}{l}{}                 & OIE2016 & WEB  & NYT  & PENN  \\ \midrule
\multicolumn{2}{l}{ClausIE~\cite{del2013clausie}}          & 58.8    & 44.9 & 29.6 & 34.6  \\
\multicolumn{2}{l}{OpenIE 4}         & 59.6    & 55.7 & 38.3 & 42.6  \\
\multicolumn{2}{l}{PropS~\cite{stanovsky2016getting}}            & 55.6    & 58.9 & 37.2 & 39.1  \\
\multicolumn{2}{l}{RnnOIE~\cite{stanovsky2018supervised}}           & 67.0    & 58.1 & 28.3 & 34.5  \\
\multicolumn{2}{l}{MAMA~\cite{wang2020language}}             & 36.6    & 54.3 & 32.9 & 33.0  \\\midrule
\multirow{3}{*}{$\quad\;$ \bf \method} & zero-shot  & 28.1         & 43.8         & 28.9       & 51.0        \\
                                  & multi-task  & 71.2         & 50.8         & 43.6       & 54.5        \\
                                  & $\quad$w/ finetune & 71.3         & 49.1         & 45.0       & 45.1        \\ \bottomrule
\end{tabular}
\caption{{Results on open information extraction.}} \label{tab:oie}
    \renewcommand\tabcolsep{7pt}
    \begin{tabular}{@{}lllllll@{}}
\toprule
   &                 & \multicolumn{1}{c}{\multirow{2}{*}{TACRED}} & \multicolumn{4}{c}{FewRel 1.0}            \\ 
&                       & \multicolumn{1}{c}{}                        & 5-1      & 5-5      & 10-1     & 10-5     \\ \midrule
\multicolumn{2}{l}{BERT$_\textrm{EM}$~\cite{soares2019matching}}     & 70.1                                        & 88.9     & -        & 82.8     & -        \\
\multicolumn{2}{l}{BERT$_\textrm{EM}$+MTB~\cite{soares2019matching}} & 71.5                                        & 90.1     & -        & 83.4     & -        \\
\multicolumn{2}{l}{DG-SpanBERT~\cite{chen2020efficient}}            & 71.5                                        & -        & -        & -        & -        \\
\multicolumn{2}{l}{BERT-PAIR~\cite{gao2019fewrel}}              &                                             & 85.7     & 89.5     & 76.8     & 81.8     \\
\multicolumn{2}{l}{NLI-DeBERTa~\cite{sainz2021label}}           & 73.9      & & & & \\
\multicolumn{2}{l}{TANL~\cite{paolini2021structured}}                   & 71.9                                        & 93.6±5.4 & 97.6±3.2 & 82.2±5.1 & 89.8±3.6 \\
\multicolumn{2}{l}{TANL (multitask)~\cite{paolini2021structured}}       & 69.1                                        & -        & -        & -        & -        \\ \midrule
\multirow{3}{*}{$\;\;$ \bf \method} & zero-shot  & 36.1         & 72.4±6.9     & 70.8±8.0   & 67.6±4.5   & 66.4±6.3     \\
                                  & multi-task  & 74.9         & 93.6±6.0     & 96.4±4.2   & 92.2±6.4   & 94.6±3.6     \\
                                  & $\quad$w/ finetune & 76.8         & 98.4±2.8     & 100±0.0    & 97.8±2.0   & 99.8±0.6      \\ \bottomrule
\end{tabular}
\label{tab:rc}
\caption{{Results on relation classification.}} \label{tab:rc}
    \renewcommand\tabcolsep{37.5pt}
    
\begin{tabular}{@{}p{2cm}lll@{}}
\toprule
\multicolumn{2}{l}{}            & Google-RE & T-Rex \\ \midrule
\multicolumn{2}{l}{LAMA~\cite{petroni2019language}} & 10.5      & 32.3  \\ 
\multicolumn{2}{l}{LAMA-Oracle~\cite{petroni2020context}} & 74.3      & 66.0  \\ \midrule
\multirow{2}{*}{$\quad\quad\quad\quad\;$\bf \method} & zero-shot  &  97.9        &  85.0\\
                                  & multi-task  &  90.3        &  71.0\\\bottomrule
\end{tabular}
\label{tab:fr}
\caption{{Results on factual probe.}} \label{tab:fr}
    \renewcommand\tabcolsep{6.9pt}
    
\begin{tabular}{@{}llllllllll@{}}
\toprule
\multicolumn{2}{l}{\multirow{2}{*}{}}   & \multicolumn{2}{c}{CoNLL04} & \multicolumn{2}{c}{ADE} & \multicolumn{2}{c}{NYT} & \multicolumn{2}{c}{ACE2005} \\ \cmidrule(l){3-10} 
\multicolumn{2}{l}{}              & Ent          & Rel          & Ent        & Rel        & Ent        & Rel        & Ent         & Rel         \\ \midrule
\multicolumn{2}{l}{SpERT~\cite{DBLP:journals/corr/abs-1909-07755}}               & 88.9         & 71.5         & 89.3       & 78.8       &            &            &             &             \\
\multicolumn{2}{l}{DyGIE~\cite{luan2019general}}               &              &              &            &            &            &            & 88.4        & 63.2        \\
\multicolumn{2}{l}{MRC4ERE~\cite{zhao-etal-2020-asking}}             & 88.9         & 71.9         &            &            &            &            & 85.5        & 62.1        \\
\multicolumn{2}{l}{RSAN~\cite{yuan2020relation}}                &              &              &            &            &            & 84.6       &             &             \\
\multicolumn{2}{l}{TANL~\cite{paolini2021structured}}                & 89.4         & 71.4         & 90.2       & 80.6       & 94.9       & 90.8       & 88.9        & 63.7        \\
\multicolumn{2}{l}{TANL (multitask)~\cite{paolini2021structured}}    & 90.3         & 70.0         & 91.2       & 83.8       & 94.7       & 90.7       & -           & -           \\ \midrule
\multirow{3}{*}{$\;\;$ \bf \method} & zero-shot  & 48.3         & 25.8         & 60.7       & 10.6       & 60.5       & 28.6       & 31.8        & 5.3         \\
                                  & multi-task  & 88.4         & 72.8         & 90.5       & 83.6       & 95.4       & 93.7       & 90.2        & 58.9        \\
                                  & $\quad$w/ finetune & 90.7         & 78.3         & 91.1       & 83.8       & 95.9       & 93.3       & 90.0        & 66.8       \\ \bottomrule
\end{tabular}
\label{tab:jer}
\caption{{Results on joint entity and relation extraction.} }
 \label{tab:jer}
    \renewcommand\tabcolsep{11.5pt}
    
\begin{tabular}{@{}llllll@{}}
\toprule
\multicolumn{2}{l}{} & CoNLL03 & OntoNotes & GENIA & ACE2005 \\ \midrule
\multicolumn{2}{l}{BERT-MRC~\cite{li2020unified}}               & 93.0    & 91.1      & -     & 86.9  \\
\multicolumn{2}{l}{BERT-MRC+DSC~\cite{li2020dice}}           & 93.3    & 92.1      &       &       \\
\multicolumn{2}{l}{Cloze-CNN~\cite{baevski2019cloze}}              & 93.5    &           &       &       \\
\multicolumn{2}{l}{GSL~\cite{athiwaratkun2020augmented}}                    & 90.7    & 90.2      &       &       \\
\multicolumn{2}{l}{BiaffineLSTM~\cite{yu2020named}}          & 93.5     & 91.3      & 80.5      & 85.4  \\
\multicolumn{2}{l}{TANL~\cite{paolini2021structured}}                   & 91.7    & 89.8      & 76.4  & 84.9  \\
\multicolumn{2}{l}{TANL (multitask)~\cite{paolini2021structured}}       & 91.7    & 89.4      & 76.4  & -     \\ \midrule
\multirow{3}{*}{$\quad$ \bf \method} & zero-shot  & 44.4         & 42.5     & 47.2       & 28.1        \\
                                  & multi-task  & 93.1         & 87.6         & 80.2       & -        \\
                                  & $\quad$w/ finetune & 93.0         & 87.8         & 80.8       & 86.9        \\ \bottomrule
\end{tabular}
\label{tab:ner}
\caption{{Results on named entity recognition.}} \label{tab:ner}

\end{table*}

\begin{table*}[]
    \centering
    \small
    \renewcommand\tabcolsep{12.5pt}
    
\begin{tabular}{@{}lllll@{}}
\toprule
\multicolumn{2}{l}{}                         & CoNLL05 WSJ     & CoNLL05 Brown & CoNLL12 \\ \midrule
\multicolumn{2}{l}{Dep and Span~\cite{li2019dependency}}             & 86.3            & 76.4          & 83.1    \\
\multicolumn{2}{l}{BERT SRL~\cite{shi2019simple}}                 & 88.8            & 82.0          & 86.5    \\
\multicolumn{2}{l}{TANL~\cite{paolini2021structured}}                     & 89.3            & 82.0          & 87.7    \\
\multicolumn{2}{l}{TANL (multitask)~\cite{paolini2021structured}}         & 89.1            & 84.1          & 87.7    \\\midrule
\multirow{2}{*}{$\quad$ \bf \method} 
                                  & multi-task  & 95.5         & 92.0         & 97.2       \\
                                  & $\quad$w/ finetune & 95.2         & 92.1         & 96.0       \\ \bottomrule
\end{tabular}
\label{tab:srl}
\caption{Results on semantic role labeling.} \label{tab:srl}
    \renewcommand\tabcolsep{8pt}
    
\begin{tabular}{@{}llllll@{}}
\toprule
\multicolumn{2}{l}{}                       & Trigger Id & Trigger Cl & Argument Id & Argument Cl \\ \midrule
\multicolumn{2}{l}{J3EE~\cite{nguyen2019one}}                   & 72.5       & 69.8       & 59.9        & 52.1        \\
\multicolumn{2}{l}{DyGIE++~\cite{wadden2019entity}}                &            & 69.7       & 55.4        & 52.5        \\
\multicolumn{2}{l}{TANL~\cite{paolini2021structured}}                   & 72.9       & 68.4       & 50.1        & 47.6        \\
\multicolumn{2}{l}{TANL (multitask)~\cite{paolini2021structured}}       & 71.8       & 68.5       & 48.5        & 48.5        \\ \midrule
\multirow{2}{*}{$\;\;$ \bf \method} 
                                  & multi-task  & 72.7         & 69.2         & 67.5      & 63.9       \\
                                  & $\quad$w/ finetune & 73.5         & 69.8         & 59.4       & 56.2        \\ \bottomrule
\end{tabular}
\label{tab:eventextractionappendixace}
\caption{{Results on event extraction (ACE2005).}} \label{tab:ee}
    \renewcommand\tabcolsep{13.4pt}
    \begin{tabular}{@{}llllll@{}}
\toprule
\multicolumn{2}{l}{}         & \multicolumn{4}{c}{CoNLL12}                                                                                            \\ \midrule
\multicolumn{2}{l}{}         & \multicolumn{1}{c}{MUC} & \multicolumn{1}{c}{B$^3$} & \multicolumn{1}{c}{CEAF$_{\phi4}$} & \multicolumn{1}{c}{Avg. F1} \\
\multicolumn{2}{l}{Higher-order c2f-coref~\cite{lee2018higherorder}} & 80.4                    & 70.8                      & 67.6                               & 73                          \\
\multicolumn{2}{l}{BERT+c2f-coref~\cite{joshi2019bert}}         & 81.4                    & 71.7                      & 68.8                               & 73.9                        \\
\multicolumn{2}{l}{CorefQA+SpanBERT~\cite{wu-etal-2020-corefqa}}       & 86.3                    & 77.6                      & 75.8                               & 79.9                        \\
\multicolumn{2}{l}{TANL~\cite{paolini2021structured}}                   & 81.0                    & 69.0                      & 68.4                               & 72.8                        \\
\multicolumn{2}{l}{TANL (multitask)~\cite{paolini2021structured}}       & 78.7                    & 65.7                      & 63.8                               & 69.4                        \\ \midrule
\multirow{2}{*}{$\quad\;$ \bf \method} 
                                  & multi-task  & 63.9        & 57.7       & 60.2       & 60.6\\
                                  & $\quad$w/ finetune & 74.9        & 71.3       & 73.1       & 73.1\\ \bottomrule
\end{tabular}
\label{tab:coref}
\caption{{Results on coreference resolution.}} \label{tab:coref}
    \renewcommand\tabcolsep{54.5pt}
    
\begin{tabular}{@{}p{2cm}ll@{}}
\toprule
\multicolumn{2}{l}{}         & MultiWOZ 2.1 \\ \midrule
\multicolumn{2}{l}{TRADE~\cite{wu2019transferable}}                  & 45.6         \\
\multicolumn{2}{l}{SimpleTOD~\cite{hosseiniasl2020simple}}              & 55.7         \\
\multicolumn{2}{l}{TANL~\cite{paolini2021structured}}                   & 50.5         \\
\multicolumn{2}{l}{TANL (multitask)~\cite{paolini2021structured}}       & 51.4         \\ \midrule
\multirow{2}{*}{$\quad\quad\quad\quad\quad\quad\;\;$\bf \method} 
                                  & multi-task  & 53.5         \\
                                  & $\quad$w/ finetune & 54.2         \\ \bottomrule
\end{tabular}
\label{tab:dst}
\caption{{Results on dialogue state tracking.}} \label{tab:dst}
    \renewcommand\tabcolsep{37.2pt}
    \begin{tabular}{@{}p{2cm}lll@{}}
\toprule
\multicolumn{2}{l}{\multirow{2}{*}{}}   &  ATIS & SNIPS \\ \midrule
\multicolumn{2}{l}{SF-ID~\cite{haihong2019novel}}            & 97.8 & 97.4  \\
\multicolumn{2}{l}{TANL~\cite{paolini2021structured}}        & 97.6 & 98.7  \\\midrule
\multirow{2}{*}{$\quad\quad\quad\quad\;\;$\bf \method} 
                                  & multi-task  & 97.3      & 97.4         \\
                                  & $\quad$w/ finetune & 97.8     & 97.3     \\ \bottomrule
\end{tabular}
\label{tab:id}
\caption{{Results on intent detection.}} \label{tab:id}

\end{table*}

\begin{table*}[]
\resizebox{\linewidth}{!}{
\begin{tabular}{@{}p{2.7cm}lp{10.5cm}p{10.5cm}@{}}
\toprule
\footnotesize
\textbf{Task}                        & \textbf{Dataset} & \textbf{Input} 
                                                        & \textbf{Output} 
                                                        \\ \midrule
Open Information Extraction          & OIE2016          & \quad oie oie2016: But for now, at least, {\color{black}Americans} are far better at {\color{black}making} {\color{black}PCs and the software that {\color{black}runs} them}.
                                                        &({\color{orange}Americans}; {\color{blue}making}; {\color{orange}PCs and the software that runs them}) ({\color{orange}PCs}; {\color{blue}runs}; {\color{orange}software})
                                                        \\
                                     & WEB              & \quad oie web: Finally {\color{black}google} {\color{black}bought} {\color{black}youtube}.
                                                        &({\color{orange}google}; {\color{blue}bought}; {\color{orange}youtube})
                                                        \\
                                     & NYT              & \quad oie nyt: Now in its 58th final, the {\color{black}United States} is {\color{black}pursuing} a 30th {\color{black}cup} title.
                                                        & ({\color{orange}United States}; {\color{blue}pursuing}; {\color{orange}cup})
                                                        \\
                                     & PENN             &\quad oie penn: {\color{black}Samsung} already {\color{black}owns} {\color{black}korea first advertising co.}, that country's largest agency.
                                                        & ( {\color{orange}Samsung}; {\color{blue}owns}; {\color{orange}korea first advertising co.} )
                                                        \\ \midrule
Relation Classification              & TACRED           & \quad rc tacred: Donald Wildmon , the founder and head of the American Family Association , is asking its members to petition Congress to end all funding for PBS . The relationship between Donald Wildmon and American Family Association is               
                                                        &  ( Donald Wildmon; {\color{blue}employee of}; American Family Association )               
                                                        \\
                                     & FewRel 1.0       & \quad rc fewrel: Boott was elected an Associate Fellow of the American Academy of Arts and Sciences in 1835 . The relationship between Boott and American Academy is              
                                                        &  ( Boott; {\color{blue}member of}; American Academy )               
                                                        \\ \midrule
Factual Probe                       & Google-RE        & \quad fp google-re: Eldon Coombe (born c 1941) is a Canadian curler from Ottawa, Canada.               
                                                        &  (Eldon Coombe; date of birth; {\color{orange}1941})               
                                                        \\
                                     & T-REx            & \quad fp t-rex: Kurt Schwertsik (born 25 June 1935, Vienna) is an Austrian contemporary composer.              
                                                        &  (Kurt Schwertsik; place of birth; {\color{orange}Vienna})              
                                                        \\ \midrule
Joint Entity and Relation Extraction & CoNLL04          & \quad jer conll04: An art exhibit at the {\color{black} Hakawati Theatre} in {\color{black} Arab} east {\color{black} Jerusalem} was a series of portraits of {\color{black} Palestinians} killed in the rebellion .
                                                        & ( {\color{orange} Hakawati Theatre}; instance of; {\color{orange} organization} ) ( {\color{orange} Arab}; instance of; {\color{orange} other} ) ( {\color{orange} Jerusalem}; instance of; {\color{orange} location} ) ( {\color{orange} Palestinians}; instance of; {\color{orange} other} ) ( {\color{orange} Hakawati Theatre}; {\color{blue} organization based in}; {\color{orange} Jerusalem} )
                                                        \\
                                     & ADE              & \quad jer ade: {\color{black} Lethal anuria} complicating high dose {\color{black} ifosfamide} chemotherapy in a breast cancer patient with an impaired renal function .
                                                        & ( {\color{orange} Lethal anuria}; instance of; {\color{orange} disease} ) ( {\color{orange} ifosfamide}; instance of; {\color{orange} drug} ) ( {\color{orange} Lethal anuria}; {\color{blue} effect}; {\color{orange} ifosfamide} )
                                                        \\
                                     & NYT              & \quad jer nyt: Mary L. Schapiro , who earlier this year became the new head of {\color{black} NASD} , was more amenable to fashioning a deal to the New York Exchange 's liking than her predecessor , {\color{black} Robert R. Glauber} .
                                                        & ( {\color{orange} NASD}; instance of; {\color{orange} organization} ) ( {\color{orange} Robert R. Glauber}; instance of; {\color{orange} human} ) ( {\color{orange} Robert R. Glauber}; {\color{blue} company}; {\color{orange} NASD} )
                                                        \\
                                     & ACE2005          & \quad jer ace2005: The {\color{black} Davao Medical Center} , a regional {\color{black} government} {\color{black} hospital} , recorded 19 deaths with {\color{black} 50} wounded .
                                                        & ( {\color{orange} Davao Medical Center}; instance of; {\color{orange} organization} ) ( {\color{orange} government}; instance of; {\color{orange} geographical entity} ) ( {\color{orange} hospital}; instance of; {\color{orange} organization} ) ( {\color{orange} 50}; instance of; {\color{orange} human} ) ( {\color{orange} hospital}; {\color{blue} part of}; {\color{orange} government} )
                                                        \\ \midrule
Named Entity Recognition             & CoNLL03          & \quad ner conll03: {\color{black} Japan} began the defence of their {\color{black} Asian Cup} title with a lucky 2-1 win against {\color{black} Syria} in a Group C championship match on Friday .
                                                        & ( {\color{orange} Japan}; instance of; {\color{orange} location} ) ( {\color{orange} Asian Cup}; instance of; {\color{orange} miscellaneous} ) ( {\color{orange} Syria}; instance of; {\color{orange} location} )
                                                        \\
                                     & OntoNotes        & \quad ner ontonotes: Relevant departments from {\color{black} Beijing Municipality} promptly activated emergency contingency plans .
                                                        & ( {\color{orange} Beijing Municipality}; instance of; {\color{orange} country city state} )
                                                        \\
                                     & GENIA            & \quad ner genia: {\color{black} Human T and B lymphocytes} demonstrate an early and transient hyperpolarization after ligand binding .
                                                        & ( {\color{orange} Human T and B lymphocytes}; instance of; {\color{orange} cell type} )
                                                        \\
                                     & ACE2005          & \quad ner ace2005: BEGALA {\color{black} Dr . Palmisano} , again , thanks for staying with {\color{black} us} through the break .
                                                        & ( {\color{orange} Dr}; instance of; {\color{orange} human} ) ( {\color{orange} Dr . Palmisano}; instance of; {\color{orange} human} ) ( {\color{orange} us}; instance of; {\color{orange} human} )
                                                        \\ \midrule
Semantic Role Labeling               & CoNLL05 WSJ      & \quad srl conll05: But while {\color{black} the New York Stock Exchange} did {\color{black} n't} [ fall ] apart Friday as the Dow Jones Industrial Average plunged 190.58 points -- most of it in the final hour -- it barely managed to stay this side of chaos .
                                                        & ( {\color{orange} the New York Stock Exchange}; instance of; {\color{orange} second argument} ) ( {\color{orange} n't}; instance of; {\color{orange} negation} )
                                                        \\
                                     & CoNLL05 Brown    & \quad srl conll05: {\color{black} His father} [ tried ] {\color{black} to make the food a topic} .
                                                        & ( {\color{orange} His father}; instance of; {\color{orange} first argument} ) ( {\color{orange} to make the food a topic}; instance of; {\color{orange} second argument} )
                                                        \\
                                     & CoNLL12          & \quad srl conll12: Dear viewers , {\color{black} the China News program} {\color{black} will} [ end ] {\color{black} here} .
                                                        & ( {\color{orange} the China News program}; instance of; {\color{orange} second argument} ) ( {\color{orange} will}; instance of; {\color{orange} modal} ) ( {\color{orange} here}; instance of; {\color{orange} location} )
                                                        \\ \midrule
Event Extraction                     & ACE2005 Trigger  & \quad ee ace2005 trg: The European Union held a {\color{black} summit} in Brussels.
                                                        & ( {\color{orange} summit}; instance of; {\color{orange} meet} )
                                                        \\
                                     & ACE2005 Argument & \quad ee ace2005 arg: The European Union held a [ summit ] in {\color{black} Brussels}.
                                                        & ( {\color{orange} Brussels}; instance of; {\color{orange} place} )
                                                        \\ \midrule
Coreference Resolution               & CoNLL12          & \quad cr conll12: And deterrents does n't work terribly well when {\color{black} an enemy} values your death more than {\color{black} his} life .
                                                        & ( {\color{orange} an enemy}; refer to; {\color{orange} his} )
                                                        \\ \midrule
Dialogue State Tracking              & MultiWOZ 2.1     & \quad dst multiwoz: [User]: I am looking for a place to to stay that has {\color{black}cheap} {\color{black}price range} it should be in a {\color{black}type} of {\color{black}hotel}. [Agent]: Okay , do you have a specific area you want to stay in? [User]: No , I just need to make sure it s cheap. Oh, and I {\color{black}need parking}. [Agent]: I found 1 cheap hotel for you that include parking. Do you like me to book it? [User]: Yes, please. {\color{black}6} {\color{black}people} {\color{black}3} nights starting on {\color{black}Tuesday}.               
                                                        &   ([User]; {\color{black}hotel area}; {\color{orange}not given}) ([User]; {\color{black}hotel book day}; {\color{orange}Tuesday}) ([User]; {\color{black}hotel book people}; {\color{orange}6}) ([User]; {\color{black}hotel book stay}; {\color{orange}3}) ([User]; {\color{black}hotel internet}; {\color{orange}not given}) ([User]; {\color{black}hotel name}; {\color{orange}not given}) ([User]; {\color{black}hotel parking}; {\color{orange}yes}) ([User]; {\color{black}hotel price range}; {\color{orange}cheap}) ([User]; {\color{black}hotel stars}; {\color{orange}not given}) ([User]; {\color{black}hotel type}; {\color{orange}hotel})
              
                                                        \\ \midrule
Intent Detection                     & ATIS             & \quad id atis: Please give me a list of all the {\color{black}flights} between Dallas and Baltimore and their {\color{black}cost}.            
                                                        &   (intent; is; {\color{orange}flight and airfare})              
                                                        \\
Intent Detection                     & SNIPS            & \quad id snips: Play the song little robin redbreast.             
                                                        &  (intent; is; {\color{orange}play music})               
                                                        \\ \bottomrule
\end{tabular}
}
\caption{{Input and output examples for every dataset.}}
\label{tab:datasetexamples}
\end{table*}

\section{Error Analysis}

We analyze the errors of {\method} 10B multi-task in recall on the CoNLL04 (JER) dataset. We specifically investigate the errors in the relation outputs. Table~\ref{tab:errmain} shows the error cases. We find that most errors are caused by minor differences between ground truth entities and predicted entities from entity outputs. For example, the predicated entity has almost the same span as the ground truth entity (e.g., ``U.S.'' and ``the U.S.''). Besides, we observe some false-positive errors that are due to the noise in the datasets. In such cases, our predictions are reasonable while they are missing due to the incompleteness of human annotations.
\begin{table*}[]
\small
\resizebox{\linewidth}{!}{
\begin{tabular}{@{}llp{12cm}p{7cm}p{7cm}@{}}
\toprule
{\bf Error type}                    & {\bf Percentage}          & {\bf Input}                                                                                                                                                                   & {\bf Ground Truth}                                        & {\bf Ours Prediction}                           \\ \midrule
\multirow{2}{*}{Close Entity}       & \multirow{2}{*}{65.3\%}   & {\sl Locations containing suitable federally owned land were listed as : Fort Wainwright annex , Fairbanks , Alaska ;}                                                        & (Fort Wainwright annex ; located in ; Fairbanks)          & (Fort Wainwright annex ; located in ; Alaska)   \\ \midrule
\multirow{2}{*}{Totally Missing}    & \multirow{2}{*}{26.4\%}   & {\sl Judith C. Toth says she returned for a fourth term in Maryland 's House of Delegates because she couldn 't find a better job .}                                          & (House of Delegates ; organization based in ; Maryland)   & (Judith C. Toth ; lives in ; Maryland)          \\ \midrule
\multirow{2}{*}{Wrong Relation}     & \multirow{2}{*}{ 4.2\%}   & {\sl After buying the shawl for \$1 , 600 , Darryl Breniser of Blue Ball , said the approximately 2-by-5 foot shawl was worth the money .}                                    & (Darryl Breniser ; lives in ; Blue Ball)                  & (Darryl Breniser ; works for ; Blue Ball)       \\ \midrule
\multirow{2}{*}{Different Focus}    & \multirow{2}{*}{ 1.7\%}   & {\sl An architect of President Nixon 's unsuccessful executive-privilege Watergate defense is a top prospect for the post of U.S. solicitor in the new Bush administration .} & ( Bush ; lives in ; U.S. )                                & ( Nixon ; lives in ; U.S. )                     \\ \bottomrule
\end{tabular}}
\caption{{\small Analysis of recall errors of \method\ on CoNLL04 joint entity and relation extraction task. For each error type, we list the percentage of missing triples caused by this particular type of error, and an example of this type of error taken from the CoNLL04 dataset.}}
\label{tab:errmain}
\end{table*}

\section{Updates from V1 to V2}

\begin{itemize}
    \item The \method multi-task results were updated in Table~\ref{tab:allres}, Table~\ref{tab:zero-shot}, Figure~\ref{fig:scaling}, Table~\ref{tab:jer}, Table~\ref{tab:srl}, Table~\ref{tab:ee}, and Table~\ref{tab:id}. We corrected the evaluation of the following tasks: joint entity and relation extraction, semantic role labeling, event extraction, and intent detection.
\end{itemize}


\end{document}